\documentclass[10pt,conference]{IEEEtran} 

\usepackage[bookmarks=true,pagebackref=true,colorlinks=true]{hyperref}
\usepackage{amsmath,amssymb,array,balance,booktabs,cite,changes,color,enumerate,float,graphicx,hyperref,multicol,multirow,siunitx,times,subcaption,mathtools,colortbl}
\usepackage[export]{adjustbox}
\usepackage{xcolor}
\usepackage[numbers,square,comma,sort&compress]{natbib}
\usepackage[font=footnotesize]{caption}
\usepackage[export]{adjustbox}
\usepackage{algorithm}
\usepackage[noend]{algpseudocode}
\usepackage{xspace}
\definecolor{myblue}{rgb}{0.44, 0.65, 0.82}
\definecolor{mygreen}{rgb}{0.032, 0.6392, 0.2039}

\newcommand{\videourl}{\url{https://sites.google.com/berkeley.edu/twirl}}

\hypersetup{colorlinks = true, breaklinks = true, citecolor = [rgb]{0,0.07843,0.45098}, urlcolor = [rgb]{0,0.07843,0.45098}, linkcolor = [rgb]{0,0.07843,0.45098}} 

\newcommand{\metabbr}{TWiRL\xspace}

\newcommand{\alert}[1]{\textbf{}}

\newcommand{\BEAS}{\begin{eqnarray*}}
\newcommand{\EEAS}{\end{eqnarray*}}
\newcommand{\BEA}{\begin{eqnarray}}
\newcommand{\EEA}{\end{eqnarray}}
\newcommand{\BEQ}{\begin{equation}}
\newcommand{\EEQ}{\end{equation}}
\newcommand{\BIT}{\begin{itemize}}
\newcommand{\EIT}{\end{itemize}}
\newcommand{\BNUM}{\begin{enumerate}}
\newcommand{\ENUM}{\end{enumerate}}
\newcommand{\BEL}[1]{\begin{equation}\label{#1}}
\newcommand{\EEL}{\end{equation}}

\newcommand{\mdp}{\mathcal{M}}
\newcommand{\state}{\mathbf{s}}
\newcommand{\goal}{\mathbf{g}}

\newcommand{\transition}{p}

\newcommand{\action}{\mathbf{a}}

\newcommand{\statespace}{\mathcal{S}}
\newcommand{\actionspace}{\mathcal{A}}

\newcommand{\policy}{\pi}

\newcommand{\rewardfunc}{r}
\newcommand{\reward}{r}

\newcommand{\discount}{\gamma}

\newcommand{\buff}{\mathcal{D}}

\newcommand{\src}{\text{src}}
\newcommand{\tar}{\text{target}}

\newcommand{\BA}{\begin{array}}
\newcommand{\EA}{\end{array}}

\DeclareMathOperator*{\expec}{\mathbb{E}}

\title{
Learning and Adapting Agile Locomotion Skills \\by Transferring Experience
}
\author{\authorblockN{Laura Smith\textsuperscript{1}, J. Chase Kew\textsuperscript{2}, Tianyu Li\textsuperscript{3}, Linda Luu\textsuperscript{2}, Xue Bin Peng\textsuperscript{4}, Sehoon Ha\textsuperscript{3}, Jie Tan\textsuperscript{2}, Sergey Levine\textsuperscript{1,2}}
\authorblockA{\textsuperscript{1}Berkeley AI Research, UC Berkeley \textsuperscript{2}Google Research \textsuperscript{3}Georgia Institute of Technology \textsuperscript{4}Simon Fraser University\\
Email: \texttt{smithlaura@berkeley.edu}}
}
\begin{document}
\setlength{\textfloatsep}{7pt}

\makeatletter
\let\@oldmaketitle\@maketitle%
\renewcommand{\@maketitle}{\@oldmaketitle%
    \centering
    \includegraphics[width=0.65\linewidth]{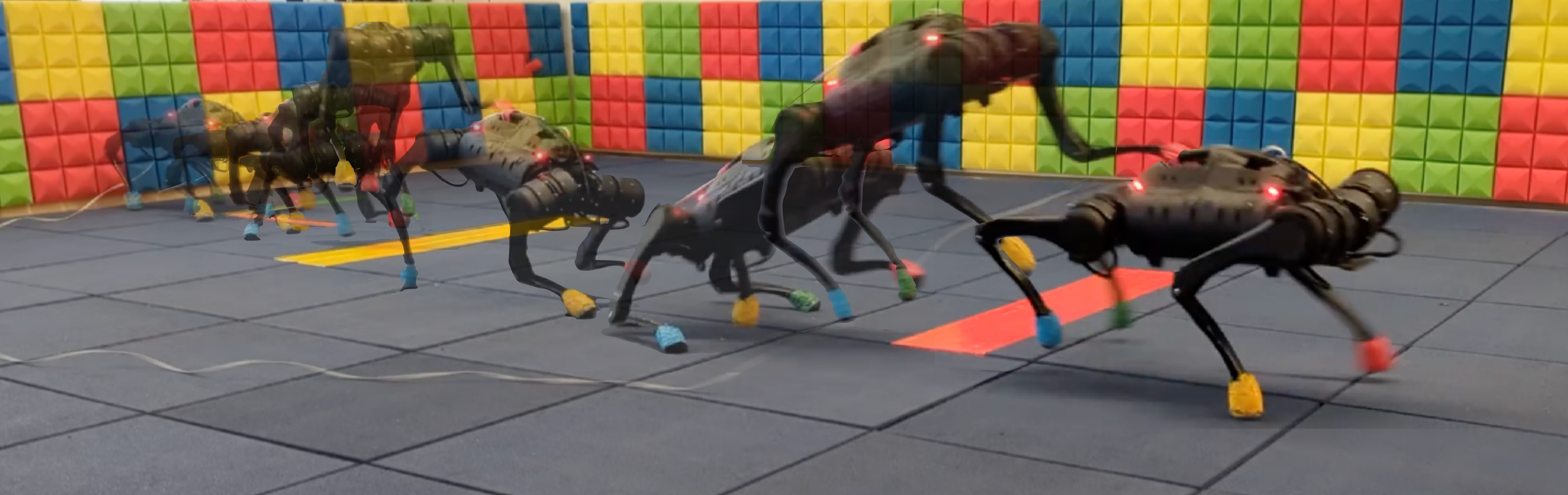}
    \includegraphics[width=0.3\linewidth]{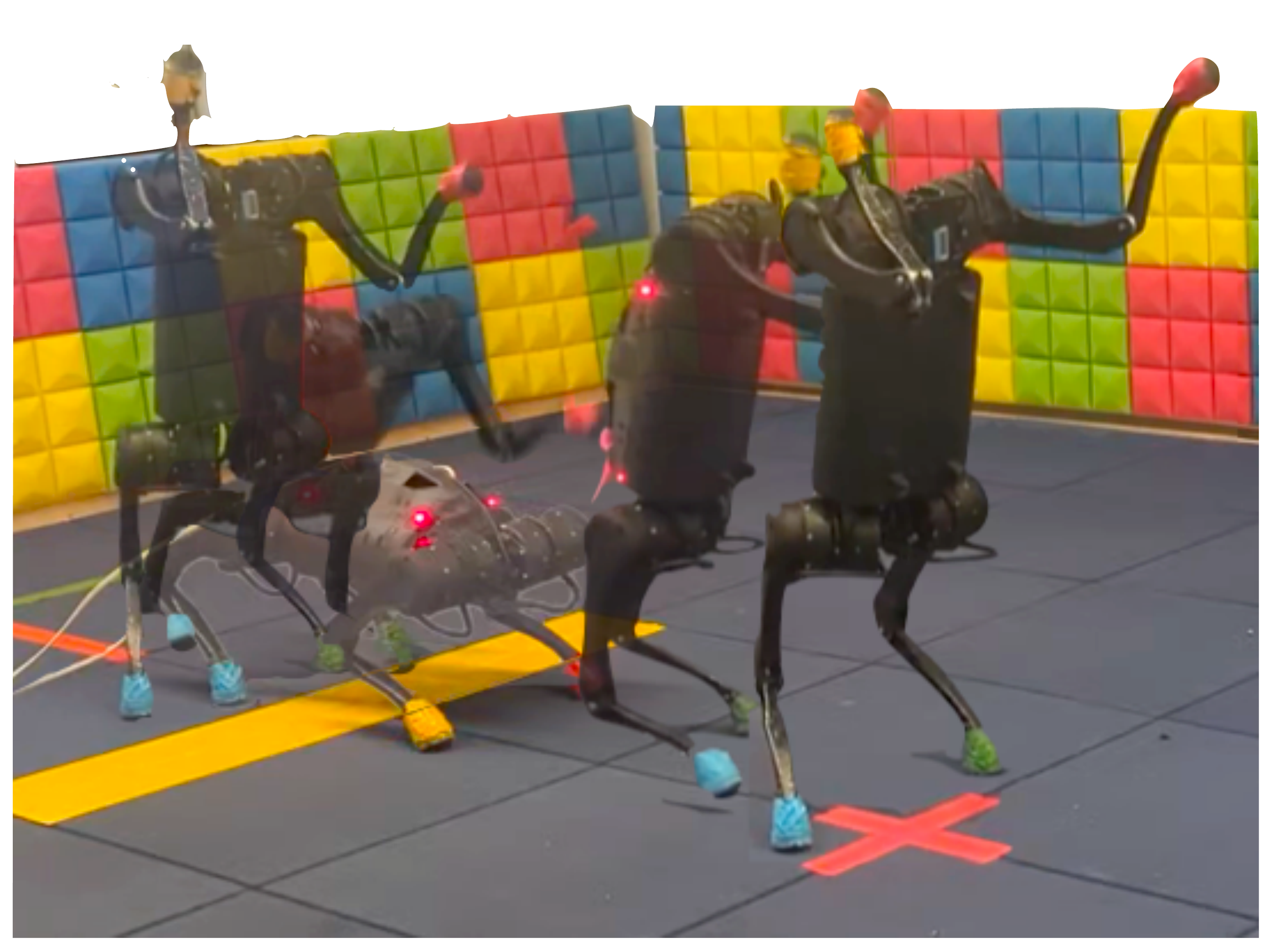}
    \captionof{figure}{\footnotesize \textbf{Agile skills} learned with our proposed method enable the A1 robot to jump repeatedly (\textbf{left}) and walk to a goal location on its hind legs (\textbf{right}).}
    \label{fig:teaser}
    \vspace{-.5cm}
}
\makeatother

\maketitle
\pagestyle{empty}

\begin{abstract}
Legged robots have enormous potential in their range of capabilities, from navigating unstructured terrains to high-speed running. However, designing robust controllers for highly agile dynamic motions remains a substantial challenge for roboticists. Reinforcement learning (RL) offers a promising data-driven approach for automatically training such controllers. However, exploration in these high-dimensional, underactuated systems remains a significant hurdle for enabling legged robots to learn performant, naturalistic, and versatile agility skills. 
We propose a framework for training complex robotic skills by transferring experience from existing controllers to jumpstart learning new tasks. To leverage controllers we can acquire in practice, we design this framework to be flexible in terms of their source---that is, the controllers may have been optimized for a different objective under different dynamics, or may require different knowledge of the surroundings---and thus may be highly suboptimal for the target task. We show that our method enables learning complex agile jumping behaviors, navigating to goal locations while walking on hind legs, and adapting to new environments. We also demonstrate that the agile behaviors learned in this way are graceful and safe enough to deploy in the real world.
(Videos\footnote{\tt\footnotesize\videourl})

\end{abstract}
\IEEEpeerreviewmaketitle

\section{Introduction}
\label{sec:intro}

\setcounter{figure}{1}
\begin{figure*}[t]
\centering
\includegraphics[width=.9\linewidth]{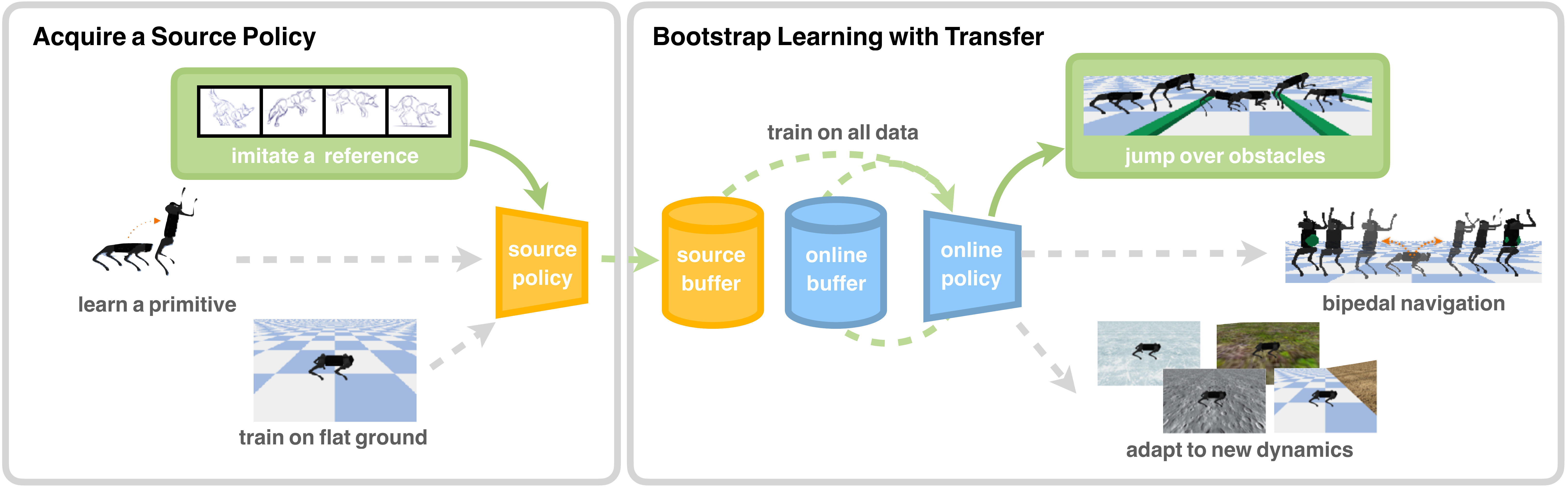}
\caption{\footnotesize Overview of three different applications of our system. We identify that several fundamental challenges in learning agile locomotion skills can be ameliorated by casting them as transfer learning problems, then applying a simple, generic method that involves a simple modification to off-the-shelf off-policy algorithms. We show that our framework is versatile---with the same method, we can \textbf{(top)} generalize a policy trained to track a reference motion of a jumping dog to learn to jump over randomly placed obstacles; \textbf{(middle)} take a policy that is trained to kick up onto the robot's hind legs to then use bipedal locomotion to navigate to randomly sampled goals; and \textbf{(bottom)} enable efficient fine-tuning in new environments. 
}
\vspace{-.5cm}
\end{figure*}
Legged animals are capable of impressive displays of agility, from mountain goats balancing on sheer inclines to search and rescue dogs bounding through rubble. Emulating such coordination could enable legged robots to go almost anywhere humans can; however, doing so is a long-standing challenge in robotics. It requires the ability to control, at high frequency, robotic systems that are high-dimensional, underactuated, and difficult to model---especially in the case of agile skills with aerial phases and unexpected contacts.
Traditional, model-based control methods have demonstrated impressive agility on legged robots, but they often require extensive knowledge of the robot and environment dynamics, as well as task-specific knowledge about the desired agile movement~\citep{Raibert1984HoppingIL, Miura1984DynamicWO, Kalakrishnan2010FastRQ, Sreenath2013EmbeddingAF, Bellicoso2018DynamicLT, Hubicki2018WalkingAR, Kim2019HighlyDQ, Chignoli2021TheMH}.

Reinforcement learning (RL), on the other hand, provides a powerful framework for autonomously acquiring robotic skills. However, learning agile locomotion policies end-to-end remains challenging for a few fundamental reasons. Foremost is that tasks with such high-dimensional systems---12 degrees of freedom for the A1 quadruped---are woefully underspecified, e.g, a robot can accomplish the simple task of moving forward in innumerable ways that are unnatural and dangerous. It is well known that even simple tasks can often require very complex, highly-shaped reward functions to elicit desired motions~\cite{Lee2019RobustRC, Lee2020LearningQL, Peng2020LearningAR, Fu2021MinimizingEC}. This underspecification problem is exacerbated by the additional complexity demanded by skills that require high speed and precision. Consider learning to jump over hurdles---the robot must first discover the coordination to launch itself in order to make any sort of meaningful progress---compared to walking, where fumbling forward still contains some useful information about how to move. That is to say, learning in this regime without extremely dense reward engineering is especially difficult because the robot can only receive learning signal by engaging in already-sophisticated, temporally coherent, and precise behaviors. So learning skills that require precise coordination of joints and contact forces entirely from scratch may be prohibitively difficult.
How might we acquire agile skills in light of these challenges?

While it may be difficult to obtain \emph{demonstrations} for a specific task (e.g. jumping to clear past variable obstructions, squeezing through tight areas, or maneuvering in extreme weather conditions), it is often possible to obtain controllers for adjacent, but simpler, settings---by training with different objectives, sensory inputs, or environments---that nonetheless contain \emph{relevant} information. Providing the robot with such relevant knowledge should significantly reduce the complexity of the learning problem by side-stepping the difficult, and sometimes insurmountable, exploration hurdle. For example, \autoref{fig:teaser} shows a task where the robot needs to walk to a specific location on its hind legs. Learning this task should be made significantly easier if the robot can already stand on its hind legs, a comparatively simpler behavior to engineer.
But these simpler behaviors are often difficult to then ``relax'' into more agile ones, especially if they are trained with heavy reward shaping to give rise to their desirable behavior.
Thus, to learn these more complex skills, we would like to somehow ``pre-train'' with the simpler or more well-shaped objectives, and then fine-tune with a minimally shaped objective that encodes the actual task we would like to solve, allowing RL to modify the motion as much as it needs to in order to achieve optimality. Adapting to new environments looks very similar---if we already have an agile skill, we might similarly prefer to fine-tune it rather than learn from scratch. In all cases, we need to answer the question: how should we train for a desired final task given another policy pre-trained on a different task, in a different environment, or with a different objective? 

The main contribution of this paper is a system for training agile robotic skills, such as jumping and walking on hind legs, facilitated by a transfer learning procedure that enables effective initialization of one skill with existing suboptimal source skills. To be broadly applicable, we present a simple, general framework for initializing skills that aims to (i) take advantage of a wide range of `source' controllers, treating them as black boxes that can be queried without requiring knowledge of their specific implementations and (ii) contend with the suboptimality of these controllers with respect to the desired task.  As further detailed in~\autoref{sec:method}, this transfer learning procedure leverages off-policy RL methods to effectively learn from the experiences generated by source controllers. We experimentally demonstrate that this provides a broadly applicable, and highly effective way to initialize skills in a target Markov decision process (MDP) with skills from other source MDPs. We then demonstrate that this can be used to provide a curriculum where simpler shaped reward and motion imitation skills can be used to bootstrap learning of highly agile behaviors, such as jumping over multiple obstacles and walking on the hind legs, and that the same exact framework can be used to transfer these skills to different environments. Finally, we show executing the learned policies enables---for the first time---a real A1 quadrupedal robot to perform consecutive running jumps across randomly placed hurdles, crossing each span of 20 cm in a \emph{single} flight phase and to walk only with its hind legs, balancing precariously on at most two of its spherical rubber feet.

\section{Related Work}
\label{sec:related}
Developing control strategies for legged robots has been a long-standing research topic in robotics. Model-based methods that leverage trajectory optimization~\citep{Bellicoso2018DynamicLT, Kalakrishnan2010FastRQ} and model-predictive control (MPC)~\cite{Hutter2016ANYmalA, Park2017BoundingCheetah, Bledt2018MITC3, Katz2019MiniCA, kuindersma2016optimization, dai2014whole} tackle this problem by using manually constructed dynamics models of the robot. These methods have shown compelling results on robust control of multiple gaits \cite{Kim2019HighlyDQ} and adaptation to complex terrains \cite{TOWR, bosworth2016robot, righetti2012quadratic}. Yet, modeling the dynamics of the robot usually requires significant expertise and engineering effort for individual problem settings. Learning-based approaches offer an appealing alternative, potentially automating the process of obtaining control policies---without requiring explicit modeling of the dynamics---and have been shown to consistently excel at enabling robust legged locomotion in simulation~\cite{Liu2018LearningBD, Lee2019ScalableMH, Peng2018DeepMimicED} and even in the real world~\cite{Kohl2004PolicyGR, Tedrake2004StochasticPG, Endo2005LearningCS, Tan2018SimtoRealLA, Haarnoja2019LearningTW, Hwangbo2019LearningAA, lee2020learning, fu2021minimizing, kumar2021rma, agarwal2022legged}.

In this work, we tackle \emph{agile} locomotion---specifically running-jump behaviors as well as bipedal navigation with a quadrupedal robot. Recently purely model-based approaches have demonstrated in the real world a variety of highly performant jumping controllers for the MIT Cheetah 2~\cite{Park2015OnlinePF, Park2021JumpingOO}, MiniCheetah~\cite{Kim2019HighlyDQ} and A1~\cite{Xie2020ALLSTEPSCL} with MPC; however, these rely on multistage planners specialized for the particular jumping motion and assume access to the dynamics model. Some work has used RL, leveraging MPC controllers to jump~\cite{Bellegarda2020RobustQJ, Margolis2021LearningTJ}. As for fully learning-based approaches,~\citet{Rudin2021CatLikeJA} learn a jumping controller focused on reorientation and landing in micro-gravity environments. In terms of comparable results to ours,~\citet{Margolis2021LearningTJ} demonstrate that using high-level velocity and foot force commands can enable a MiniCheetah to continuously hop over randomly placed gaps. When transferred to the real world, the robot is able to cross 26cm gaps in two hops (the first to center its body over the gap and transfer its front feet across and the second to then transfer its back feet). In contrast, we utilize low-level PD target actions and can directly apply our method to learn other motions (such as bipedal locomotion). We further demonstrate an A1 robot clearing 20cm gaps in the real world in \emph{one} leap, moving its entire body over the gap in one highly dynamic motion. As for bipedalism with quadrupedal robots,~\citet{Vollenweider2022AdvancedST}, who use adversarial motion priors to get an ANYmal to balance on its hind legs with large wheels as feet and do not then demonstrate walking. \citet{Yu2022MultiModalLL} demonstrate bipedal walking with a quadrupedal robot; however, they add a supporting stick to both back feet, making it possible to get up quasi-statically and still have 4 points of contact while walking, so the policy does not require a high degree of agility. \citet{Fuchioka2022OPTMimicIO} use RL to imitate a reference motion of bipedal stepping generated with trajectory optimization on a Solo 8 robot~\cite{Grimminger2019AnOT}. As we discuss below, our work is complementary to such imitation-based policies since our approach is to use their data to learn more complex tasks.

As mentioned in~\autoref{sec:intro}, the primary challenges in learning high dimensional underactuated robotic systems are task specification and exploration. A popular way to address these two challenges is imitation learning~\cite{schaal1997learning, ng2000irl, abbeel2004irl, argall2009il}, where the agent is trained to copy an expert. In locomotion, this principle often manifests as tracking reference motions~\cite{2018-TOG-deepMimic, 2018-TOG-SFV, bergamin2019drecon, fussell2021supertrack, won2020scalable, RoboImitationPeng20, Fuchioka2022OPTMimicIO}. The reference motion in these cases specifies the task completely and provides very dense feedback, addressing the exploration problem. The drawback of imitation-based approaches, though, is that in exchange for solving the exploration challenge, they are restricted to the ``expert" or reference motion as the policy's objective is to track it. That is, they are not \emph{directly} applicable to solve downstream tasks or be adapted to situations in which the expert performs sub-optimally~\cite{Peng2021AMPAM, Escontrela2022AdversarialMP}. \citet{Bogdanovic2021ModelfreeRL} make a similar observation that motion-tracking can be brittle to environment uncertainties, and so first use an imitation objective to learn a policy that is then fine-tuned with randomization on an actual task objective. Our work specifically explores initializing the replay buffer as an effective avenue for cross-task information transfer, with the benefit of being able to utilize data from existing policies (with different action and observation spaces) without having to first train with a motion-tracking objective.

When reference motions are unavailable, another option is heavy reward engineering, with or without a curriculum. \citet{Rudin2021LearningTW} report using a 9-term reward function to learn a velocity-conditioned locomotion task, with manually-tuned weights to balance the various reward terms. This method can also yield impressive, more agile controllers, with the investment of sufficient effort in tuning the parameters. \citet{margolisyang2022rapid} use a similar setup with an adaptive curriculum on the velocity command to achieve a 3.9 m/s controller for the MiniCheetah. Several works show promising results using automatic curricula without the need to encode extensive domain knowledge, but they have yet to be demonstrated outside of simulation~\cite{Iscen2020LearningAL, Tang2020LearningAL}.  
These are all valid ways to obtain controllers, and we in fact leverage some of them in our own work. Our method is concerned with bootstrapping \emph{from} existing policies to accomplish more difficult tasks. Therefore, any of these existing training methods can be seen as complementary to our method. 

Our method applies the concept of transfer using a general method in order to address the aforementioned challenges. One facet of transfer very commonly studied in the realm of locomotion is that of shifts in dynamics (e.g., sim-to-real, different types of terrains). Domain randomization is widely used as a method for achieving generalizable policies by training them to do well on average under sufficiently diverse conditions~\citep{Cutler2014ReinforcementLW, Sadeghi2017CAD2RLRS, Rajeswaran2017EPOptLR, Tobin2017DomainRF, Peng2018SimtoRealTO, Yu2019SimtoRealTF, hwangbo2019learning, Xie2019LearningLS, Yu2020LearningFA, Peng2020LearningAR} so as to ensure that the resulting policy can be adapted to variations of its training environment. Meta-RL is another line of work that achieves few-shot adaptation by directly optimizing the adaptation procedure~\citep{Wang2017LearningTR, Finn2017ModelAgnosticMF, Rakelly2019EfficientOM, Song2020RapidlyAL}. These approaches all require some sort of distribution of training conditions, and an assumption that the test conditions will be likely under this distribution. In order to be able to leverage existing controllers, we cannot make such assumptions on the training distribution or origins of our source policies. As such, we use the \emph{experience}, which we can get by running the source policies in our target domain, to bootstrap learning from\added{.}

While there is a breadth of work in initializing agents with prior data to bootstrap from~\cite{Hester2017DeepQF, Rajeswaran2017LearningCD, Rudner2022OnPI, Nair2017OvercomingEI, playbackcontrol, LearningFromDemos}, these methods often make the assumption that the source policy is optimal and \emph{constrain} the agent to behave similarly to the demonstration data. The nature of these problems is that the demonstrations are assumed to be optimal with respect to the task, and the RL is needed in order to learn a policy that is robust to variations. Most similar in spirit to our work is DDPGfD~\cite{Vecerk2017LeveragingDF} and~\citet{Nair2017OvercomingEI}, in that they leverage off-policy RL to use a pre-collected dataset to overcome exploration challenges. They don't, however, consider changes to the MDP like we do to apply this principle to learning agile locomotion. \citet{Xie2021LifelongRR} employ a similar methodology to ours in the continual learning setting while leveraging functional access to the new task's reward to relabel past experiences and filtering to only continue learning on transitions that are similar to the online data. They demonstrate results on a sequence of manipulation tasks with end-effector control at 5Hz, we instead study the ability to apply this principle to very difficult locomotion skills with low-level control at 20Hz.

\section{Preliminaries}

\label{sec:preliminaries}
We frame learning a locomotion skill as a Markov decision process (MDP)~\citep{MDP} $\mdp$, defined as a tuple $(\statespace, \actionspace, \discount, \transition, \rewardfunc)$, describing an agent situated in an environment with state space $\statespace$ and action space $\actionspace$, whose dynamics are governed by a transition function $\transition(\state'\mid\state, \action)$. Given a reward function $\rewardfunc(\state, \action)$ and discount factor $\discount$, the RL objective is to learn a policy $\policy$ that maximizes the agent's expected discounted return: $\expec\left[\sum_{t=0}^\infty \gamma^t \rewardfunc(\state_t, \action_t) \right]$.

\paragraph{Transfer learning} We are interested in the setting where we have access to a policy $\pi_\src$ trained in a source MDP $\mdp_\src$, which is structurally related to the desired target MDP $\mdp$. For example, $\mdp_\src$ might have a different reward function (e.g., imitating a jumping motion rather than actually clearing an obstacle), a different state space (e.g., $\mdp$ includes a desired target to walk to, but $\mdp_\src$ does not), or different dynamics (e.g., $\mdp_\src$ involves jumping on flat ground, but $\mdp$ requires jumping on sloping ground). This includes both curriculum learning problems, where the policy might be difficult to learn in $\mdp$ directly but easier when appropriately leveraging $\pi_\src$, and problems where we might want to accelerate learning of $\policy$ by leveraging a $\pi_\src$ that is already performant in some other setting. In all of these cases, we would like to use $\pi_\src$ to aid in learning $\policy$ in the target MDP $\mdp$, but it is not obvious how this should be done. As we will discuss and illustrate in our experiments, na\"{i}vely initializing $\policy$ from $\pi_\src$ is not always possible, and often is not the best choice. 

\paragraph{Off-policy RL} While there are many algorithms for optimizing this objective, we specifically consider \emph{off-policy} methods, as they can in principle incorporate data from other sources---like those collected by hand-designed, simpler, or more narrow policies. Many off-policy deep RL algorithms for continuous control~\cite{Haarnoja2018SoftAO, Xu2020DeepDP} fit a critic using a dataset of experiences to estimate its performance and then update the policy accordingly. This dataset of experiences, or replay buffer $\buff$, is usually composed of the robot's own past experiences while learning in a particular MDP. We will discuss how such methods can be used as part of a transfer learning procedure in several different ways, and present a specific approach that we find to work well across a variety of agile quadrupedal control problems.
\section{Transfer Learning for Agile Skills}
\label{sec:method}
We present ~\metabbr: \textbf{T}ransferring \textbf{Wi}th off-policy \textbf{RL}, a simple yet effective model-free RL framework for learning complex locomotion skills. In this section, we argue that transfer learning provides a very powerful tool for addressing the exploration challenges outlined in~\autoref{sec:intro} and thus makes it feasible to learn highly agile robotic locomotion skills. We then detail a practical method for facilitating this transfer through simple modifications to off-policy RL.

\subsection{Bootstrapping Capable and Flexible Policies}
\label{sec:abstractapps}
Exploration is particularly daunting for agile locomotion due to the complex coordination and precision required combined with the aforementioned challenges in task specification and high-dimensional control. We propose a method to leverage existing controllers to help overcome this challenge; however, it is difficult to devise a method that can adapt policies trained in MDPs that can differ in various ways from the target MDP. First, while they may be useful in providing examples of decent behaviors, out-of-the-box policies are likely highly suboptimal for the target task. As agile skills require precision, the robot should be able to pick out \emph{only} what is useful, adapting or building upon it to accomplish the target task without inheriting idiosyncrasies or irrelevant behaviors. A natural idea is to simply \emph{fine-tune} policies for the new task, but this may not always be possible, especially in the applications we consider. In the case of transferring policies trained with different objectives, the observation space will often change \emph{along with} the reward function as the robot may need additional information to solve a task. For example, we may want to use a blind (trained only from proprioception) policy that has focused on learning a difficult control problem to aid in learning a task that requires knowledge of its surroundings. Another case in which we cannot just straightforwardly fine-tune a policy is if we want to bootstrap learning from policies that were not learned via RL, e.g., traditional model-based methods~\citep{Kim2019HighlyDQ, Bellicoso2018DynamicLT, TOWR, dai2014whole}. To handle all these cases, a key desideratum of our system is it should be agnostic to \emph{how} the source policy is obtained. 

\subsection{\metabbr: Approach Overview}
\label{sec:algorithm}
\begin{algorithm}[t]
  	\caption{\footnotesize \metabbr pseudocode}
  	\label{alg:finetune}
  	\begin{algorithmic}[1]{
\footnotesize
          \Require Source policy $\policy_\src$, MDP $\mdp_\src$ , MDP $\mdp_\tar$ of the desired task
  	 \State Initialize: Source policy replay buffer $\buff_\src$, online replay buffer $\buff_\tar$, sampling ratio $\phi$, policy parameters $\theta$
    \item[]
        
        \State {{\textsc{// Obtain data from $\policy_\src$}}}
        \If{$\buff_\src$ is empty}
        \Repeat
            \State Collect data with $\policy_\src(\action \mid \state)$.
      	\State Add experience to source buffer $\buff_\src \leftarrow (\state, \action, \state', \reward)$
  	\Until{desired}
        \EndIf
    \item[]
        \State {{\textsc{// Train on all data using high UTD off-policy RL}}}
  	\Repeat
      	\State Collect data with $\policy_\theta(\action \mid \state)$.
      	\State Add experience to online buffer $\buff \leftarrow (\state, \action, \state', \reward)$
            \State Construct $\beta_\src$ sampling $\phi\cdot$batch size from $\buff_\src$
            \State Construct $\beta_\tar$ sampling $(1-\phi)\cdot$batch size from $\buff$
            \State Update $\theta$ using $\beta_\src\cup\beta_\tar$ 
  	\Until{converged}
        }
  	\end{algorithmic}
\end{algorithm}

Our key insight is that transferring \emph{experience} provides a general avenue through which to bias the robot's exploration without constraining it.
The only assumption we make is the ability to execute actions from the source policy in $\mdp$ and record the observations, actions, and rewards. The question now is how we can effectively leverage this possibly suboptimal data to bootstrap learning. Since off-policy RL uses dynamic programming to propagate the information, it can in principle connect its online experience to relevant parts of the source policy data and quickly coopt effective strategies exhibited by the source policy. We find empirically that incorporating the data from $\policy_\src$ as transitions in the replay buffer for training $\policy$ is a surprisingly effective strategy when implemented with a couple of key design decisions we will introduce as we now describe the overall procedure. 

Our method (summarized in~\autoref{alg:finetune}) is formulated as follows:  We start by constructing a dataset $\buff_\src$ by rolling out the source policy $\policy_\src$ in $\mdp$.
We then train our policy $\policy_\theta$ using off-policy RL with data sampled from both $\buff_\src$ and the new policy's replay buffer $\buff$. We sample at a constant ratio $\phi$ from $\buff_\src$ (and $(1-\phi)$ from $\buff$) to remain robust to the size of $\buff_\src$. We find through extensive experiments (please refer to the supplementary material) that this strategy and setting $\phi=0.5$ strikes a good balance between reliability and performance. Finally, the last component we find to be important (again, please refer to the supplementary material for ablations) is using a high \emph{update-to-data ratio}, the ratio between policy updates (\autoref{alg:finetune}, line 14) and data collection (line 10). This was also observed by~\citet{Vecerk2017LeveragingDF}, who used off-policy learning to incorporate human demonstrations to overcome sparse reward problems. They introduced $L_2$ regularization on the policy weights to accommodate this, whereas we find that the stochastic regularization used in~\cite{Hiraoka2021DropoutQF} sufficient. Intuitively, taking more gradient updates may be important for effectively processing and incorporating the source policy data.
\section{Learning Agile Locomotion Tasks with \metabbr}
\label{sec:experimentsetup}

\metabbr is a robot learning system for acquiring complex locomotion skills by leveraging experience from related tasks or environments. 
In this section, we introduce concrete applications and how our framework can be applied to solve them.

\subsection{Setup Details}
\label{sec:MDPsetup}
We use the A1 robot from Unitree as our platform and build our simulation using PyBullet~\cite{coumans2021}. We define the state $\state_t$ to be a three-step history of the following features: root orientation, joint angles, base displacement, and previous actions. Our actions $\action_t$ are the target joint angles for the 12 actuated degrees-of-freedom. We do not adopt any additional engineered architectures for actions~\citep{Kohl2004PolicyGR, Iscen2018PoliciesMT, Lee2020LearningQL, Yang2022SafeRL} due to the generality of the agile motions that we aim to learn. The sensing-actuation control loop runs at a frequency of 20Hz. 

The off-policy RL algorithm we use in our approach (outlined in~\autoref{sec:algorithm}) is a state-of-the-art off-policy actor-critic algorithm DroQ~\cite{Hiraoka2021DropoutQF}, a variant of SAC~\citep{Haarnoja2018SoftAO} that incorporates dropout regularization and layer normalization~\cite{Ba2016LayerN}. We use the open-sourced JAX~\cite{jax2018github} implementation from~\citet{Smith2022AWI}. We report the average return and standard deviation across 3 runs unless otherwise stated.

\subsection{Applications} \label{sec:applications}

\subsubsection{Reward curriculum}
\label{sec:jumpingsetup}
While jumping over obstacles can drastically improve the mobility of a quadrupedal robot in cluttered environments, it is a particularly difficult skill for robots to master~\citep{Iscen2020LearningAL,agarwal2022legged}. For a successful jump, the robot must not only (i) reach the required speed to generate sufficient momentum to launch itself, but also (ii) dynamically adjust its step size depending on how soon it needs to jump. Then, while landing, the robot must learn to effectively (iii) leverage its high-bandwidth actuators to properly stabilize itself against high-impulse contact forces. In contrast, the controllers most commonly studied for legged robots are quasi-static, with the robot being grounded by at least one foot at all times. 

We design our jumping task so that the robot must contend with these three challenges by requiring the robot to jump consecutively over 20cm wide, 1cm tall `hurdles' that are spaced randomly between 1.6 and 2.6 meters apart. As such, the robot should run up to each hurdle to jump over it, then land while continuing to run in order to execute the next jump at the right time. We define the reward function for this task with minimal shaping:
\begin{equation}
    \rewardfunc(\state, \action)^\text{jump} = \reward^\text{rv}(\state, \action)  + \reward^\text{o}(\state, \action) + \reward^\text{jv}(\state, \action) + \reward^\text{p}(\state, \action),
\end{equation}
where root velocity $r^\text{rv}$ and progress $r^\text{p}$ rewards encourage the robot to move forward, $r^\text{o}$ penalizes undesirable orientations, and $r^\text{jv}$ penalizes joint velocities. Please refer to the supplementary material for exact definitions. We terminate the episode if the robot collides with the hurdle.
In order to learn the task, in addition to the features listed in~\autoref{sec:MDPsetup}, the observation space also contains a displacement vector between the robot's root position and the center of the hurdle, projected onto the ground plane. 
\begin{figure}[t]
\centering
\includegraphics[width=.9\linewidth]{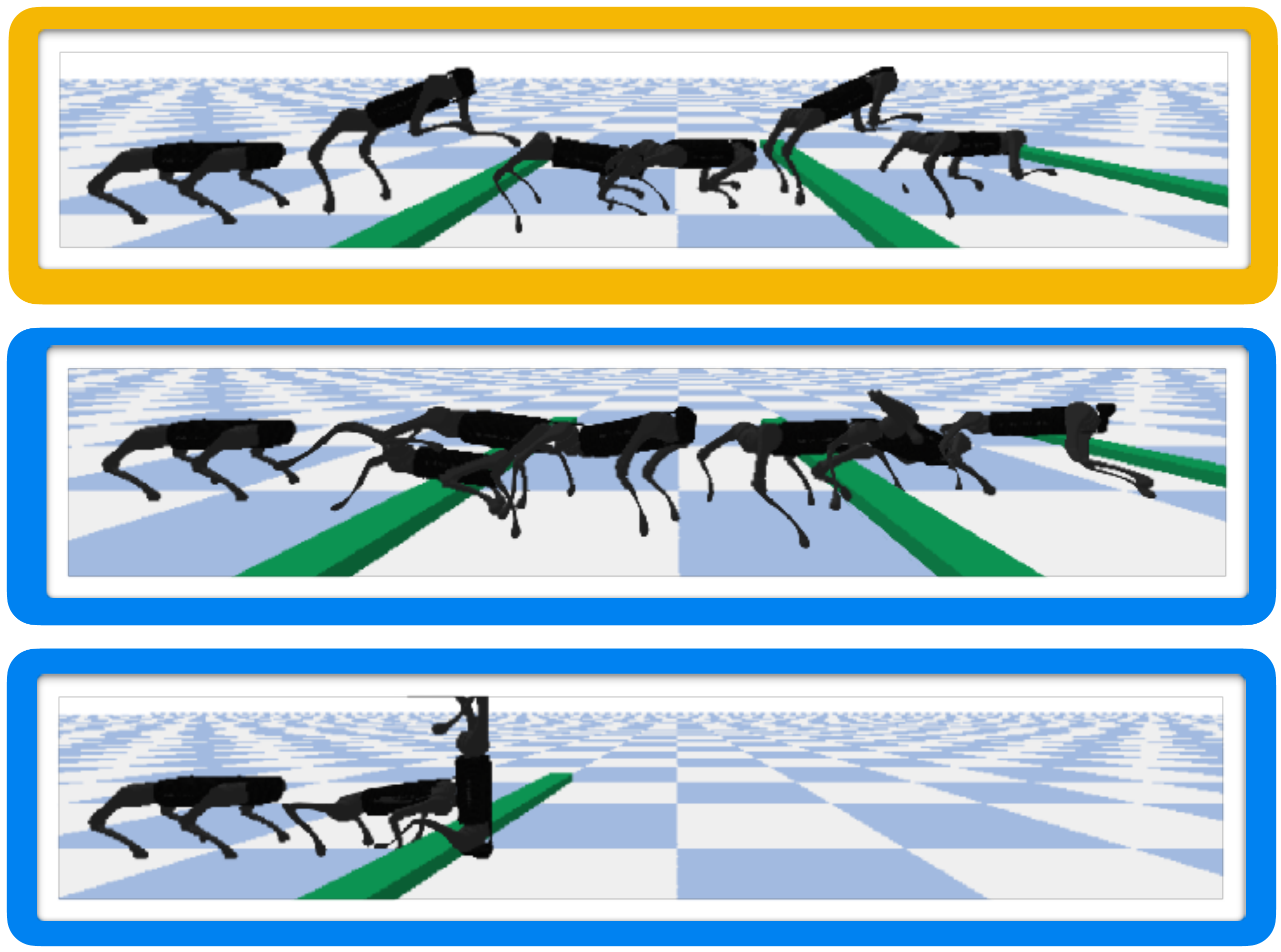} 
\caption{\footnotesize Examples of our policy (outlined in yellow), which incorporates both online training and data from a motion imitation policy, compared to two policies (outlined in blue) trained from scratch with the same reward function. While na\"ively optimizing for the task either exploits the simulator to learn an unnatural motion (middle) or fails completely (bottom), the policy trained by incorporating prior data exhibits a graceful jump.}
\label{fig:jumping-qualitative}
\end{figure}

As we see in~\autoref{fig:jumping-qualitative}, training directly with the task reward in this case has two failure modes. First, we observe reward exploitation~\cite{amodei2016concrete,implicit-preferences, reward-side-effects}, where the robot steps over the hurdles unnaturally (please see the supplementary material to see videos of the behavior). In other runs, the robot greedily tried to move forward without hitting the hurdle by somersaulting (which was ultimately unsuccessful) and could not escape this local minimum.
So, while this task is not well-suited to learning from scratch with RL, can this problem be solved by incorporating data from a policy that we~\emph{can} easily acquire? Prior work has come up with methods for imitating reference motions, enabling agents to learn naturalistic behaviors with shaped reward functions~\cite{2018-TOG-deepMimic, 2018-TOG-SFV, bergamin2019drecon, fussell2021supertrack, won2020scalable, RoboImitationPeng20}. Since these policies follow a fixed pattern (as prescribed by the reference motion), it is not robust to the variations required by the task we define of jumping over randomly placed hurdles. Nonetheless, we expect them to be useful to learn this more general task. As \metabbr is agnostic to the specific method used to produce the source policy, for our experiments, we choose to use the system proposed by~\citet{Peng2020LearningAR} to learn a policy that imitates a leaping motion taken from the dataset provided by~\citet{zhang2018mode}. Here, we collect 50k samples by rolling out this policy (trained in $\mdp_\src$) in $\mdp_\tar$ to construct $\buff_\src$. Notably, \emph{we do not initialize the task policy with the source policy is because of the mismatch in observations}---the motion imitation policy takes the fixed referenced trajectory as part of the input, while the task policy includes the distance to the next hurdle in observation for adapting to random hurdle locations. In Section \ref{sec:experimentresults}, we demonstrate that the jumping task can be solved perfectly by incorporating this data via~\metabbr.

\subsubsection{Task curriculum}
\label{sec:bipedsetup}
\begin{figure}[t]
\centering
\includegraphics[width=.65\linewidth]{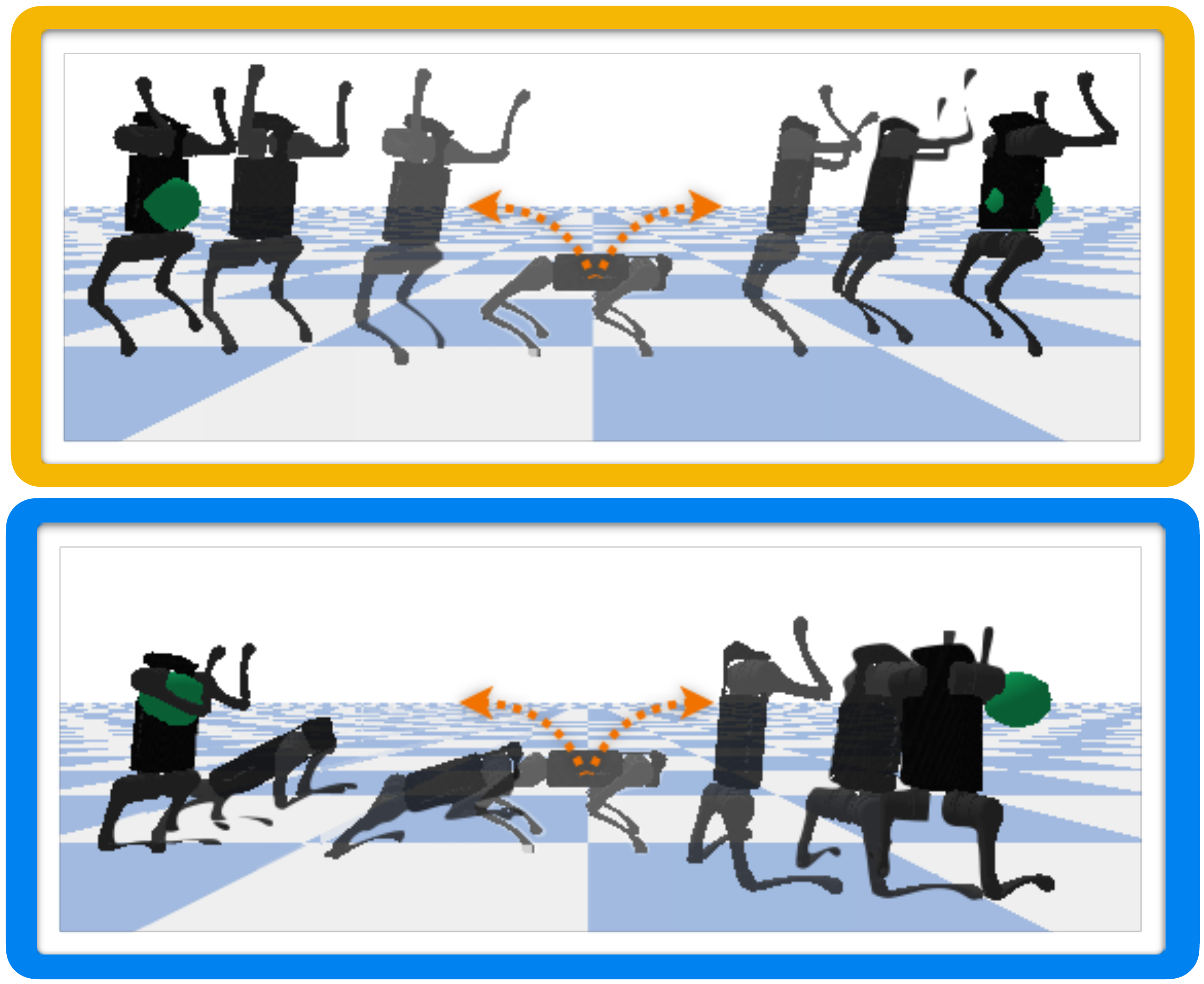} 
\caption{\footnotesize Examples of our policy (outlined in yellow), trained with data from a robot that can already stand on its hind legs, compared to a baseline policy (blue) trained from scratch. Without this added bias, the baseline policy learns to scoot toward the goal on its knees. Our policy gracefully kicks up to standing and navigates to the goal on 2 legs.}
\label{fig:bipedal-qualitative}
\end{figure}

Bipedal walking requires less space and enables walking through narrow spaces. It can also accomplish tasks that are difficult for quadrupedal robots, such as climbing stairs and doing manipulation work with the front legs. However, the advantages also come with extra difficulties. One big challenge is balancing---the robot is inherently unstable on its hind legs, and maintaining balance while walking requires continuous adjustments to the body's center of mass, which is achieved through coordinated and precise control of the motors. Note that, unlike dedicated bipedal robots, the A1 has small rounded feet and a body shape that makes balancing on the hind legs more difficult.

In our experiments, we define a goal-conditioned navigation task in which the robot must acquire the agility to get up on its hind legs and walk to a desired location while maintaining balance. We randomly sample these goals along a circle with a random radius, so as to train the robot to walk to a wide variety of goal locations. We again define a reward function with minimal shaping:
\begin{equation}
    \reward(\state, \action) = r^\text{s}(\state, \action) \cdot (1 + r^\text{f}(\state, \action) + r^\text{d}(\state, \action)).
\end{equation}
The `stand' reward $r^\text{s}_t$ encourages the robot's forward vector to be perpendicular to the ground plane and to maximize the heights of its front feet. The `facing' reward $r^\text{f}_t$ encourages the robot's belly to be pointed toward the goal, and the distance reward $r^\text{d}_t$ gives the robot a constant bonus if its distance to the goal decreased during that step. For full details, please refer to the supplementary materials. The reward function encourages the robot to move towards the goal while maintaining the bipedal standing pose, as $r^\text{s}_t$ acts as a gating function. 
For the task, the policy also receives the displacement vector, projected onto the ground plane, between itself and the goal.

As we show in~\autoref{fig:bipedal-qualitative}, training directly with the task reward again does not lead to successful learning. However, if we have a policy that has first learned just to stand, we can leverage experience from this policy with our method to make the bipedal navigation task practical to learn: We trained our source policy $\policy_\src$ with $\reward(\state, \action) = r^\text{s}(\state, \action)$ just to get it to learn to stand on its hind legs. Then, after the robot has mastered getting up, it can move on to learning how to walk to the goal. Similarly to the jumping task, we collect 50k samples to populate $\buff_\src$ by rolling out this policy in $\mdp_\tar$.
\begin{figure}[t]
\centering
\includegraphics[width=.75\linewidth]{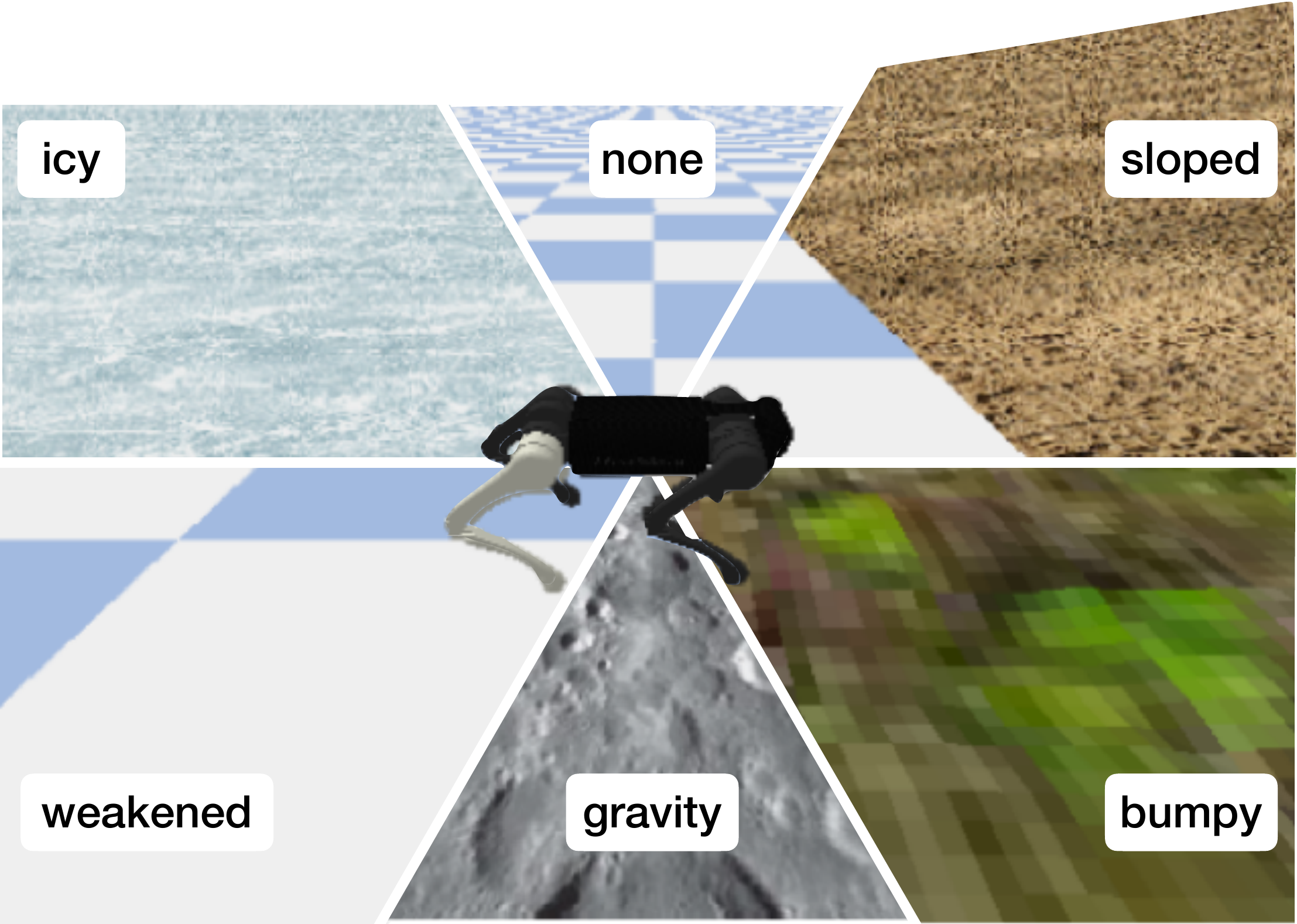}
\captionof{figure}{\footnotesize Illustrating the diverse environments to which we adapt the source policy (trained in the environment labeled `none' to indicate no modification). In clockwise order: the default, non-randomized environment; a sloped terrain; bumpy terrain; a low-gravity environment; a stochastic environment simulating motor weakening; a simulated ice rink.}
\label{fig:sim-domains}
\end{figure}

\subsubsection{Adapting skills to different environments}
As the real world is complex and unpredictable, robots deployed in the wild will inevitably encounter situations for which they are not yet prepared and need to adapt their motor skills accordingly. A benefit of our framework which is illustrated in the two previous use cases is that it should allow us to learn skills with general, non-environment-specific reward functions, making them amenable to adapt to other environments. In our last example, we test our ability to adapt these skills to different environments using~\metabbr. To do so, we set up a set of ``sim-to-sim'' transfer experiments, where we pre-train $\policy_\src$ in the same simulated environment and then transfer the policy in a variety of \emph{other} simulated environments whose dynamics $\transition$ differ from the pre-training setting. Specifically, we transfer under the conditions depicted in~\autoref{fig:sim-domains} (for exact details please see the supplementary material).
This setting is distinct from the other two in that it is possible in this case to \emph{also} transfer the policy network since only the dynamics are changing. Therefore, in these experiments, we will additionally study whether transferring the policy is helpful.
\section{Results}
\label{sec:experimentresults}
\begin{figure}[t]
\centering
\begin{subfigure}{\linewidth}
\centering
\includegraphics[width=\linewidth]{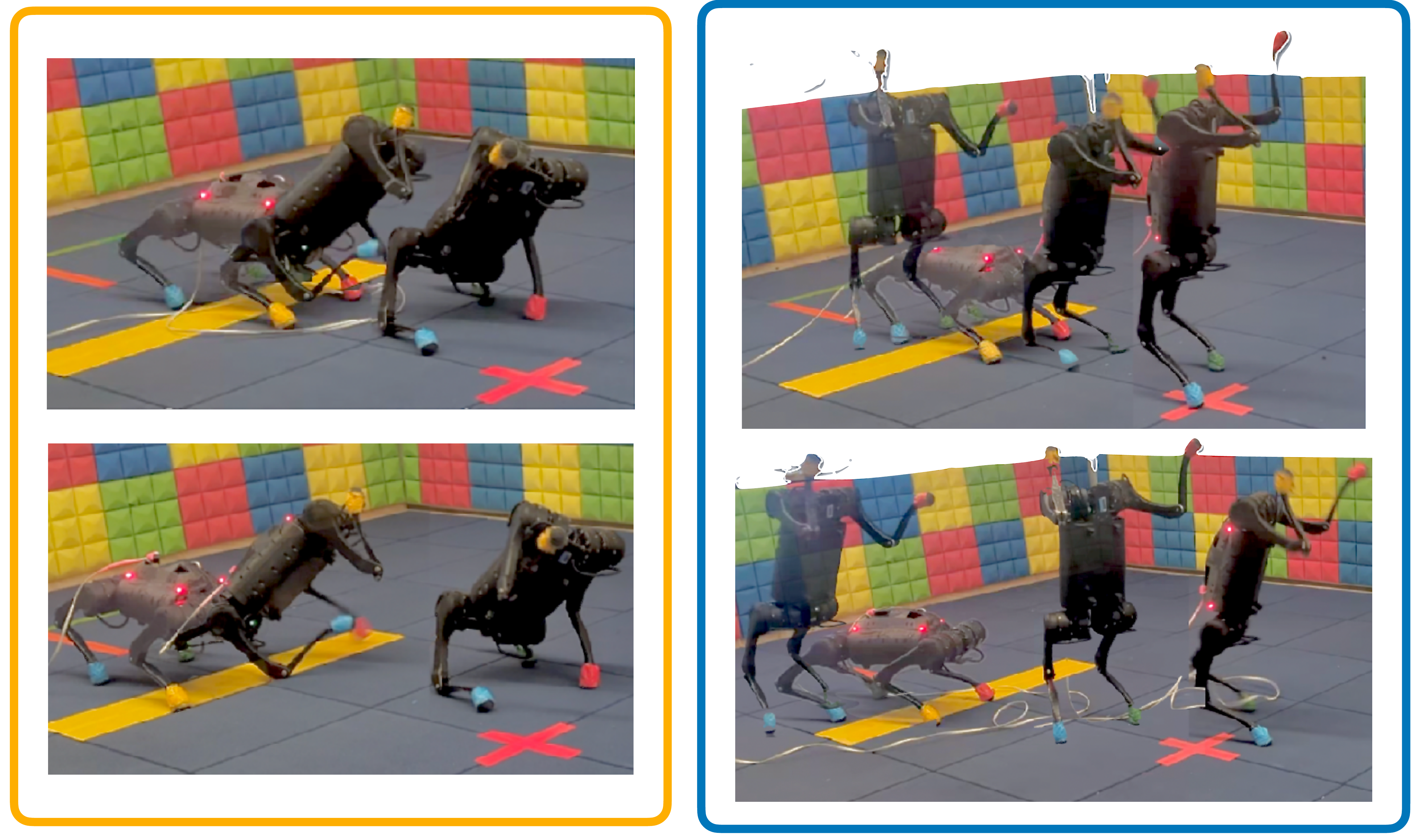}
\caption{\footnotesize Examples a policy learned from scratch (left, yellow) and learned by~\metabbr (right, blue) rolled out in the real world. Videos of all evaluation trials can be found in the supplementary material.}
\label{fig:real-biped-qualitative}
\end{subfigure}%
\vspace{.25cm}
\begin{subfigure}{\linewidth}
\centering
\begin{tabular}{c | c | c}
\toprule
\rowcolor[HTML]{ededed}
 Metric & RL from Scratch & \metabbr\\
\midrule
Standing Time & 0.00 s & 12.12 s \\
\midrule
Min. Distance &  0.49 m & 0.14 m  \\
\midrule
Return & 589 & 3127\\
\bottomrule
\end{tabular}
\caption{\footnotesize We report the time the robot remained standing, the minimum distance the robot got to the goal, and the return as defined by our task.}
\label{tab:real-biped-quantitative}
\end{subfigure}
\caption{\footnotesize Comparison of our bipedal navigation policy to the baseline trained from scratch evaluated in the real world. Over 5 trials, our policy spent an average of 12 seconds upright and successfully approached the goal, whereas the baseline consistently scooted on its knees and never stood upright.}
\label{fig:real-biped-combined}
\end{figure}

We now present our experimental analysis of~\metabbr. As a baseline, we evaluate directly applying RL to solve the tasks introduced in~\autoref{sec:experimentsetup} from scratch. We then demonstrate how these tasks can be addressed more effectively by casting them as transfer problems for which we can readily acquire source policies---we show that incorporating off-policy data can (i) overcome exploration challenges, and (ii) bias the learning algorithm toward acquiring locomotion skills that are agile, robust, and safe for execution on physical hardware. Our experiments aim to answer the following concrete questions:
\begin{enumerate}
    \item Can current RL methods handle the challenges of learning agile locomotion skills?
    \item Can incorporating data from other policies using~\metabbr enable the robot to learn these challenging tasks? 
    \item Can~\metabbr policies be deployed in the real world?
    \item Can we apply~\metabbr to adapt to different dynamics?
\end{enumerate}
\begin{figure}[t]
\centering
\includegraphics[width=.49\linewidth]{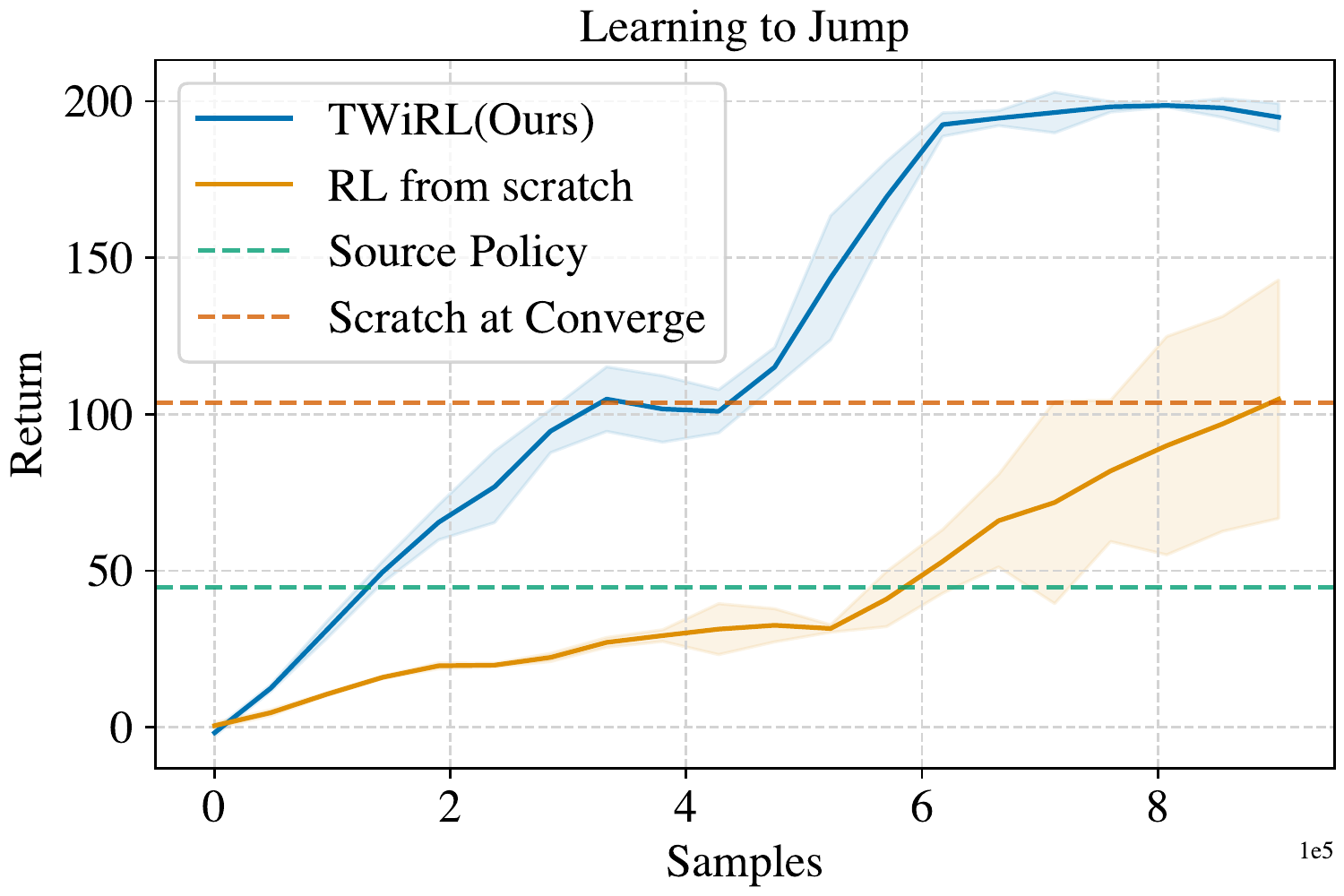}
\includegraphics[width=.49\linewidth]{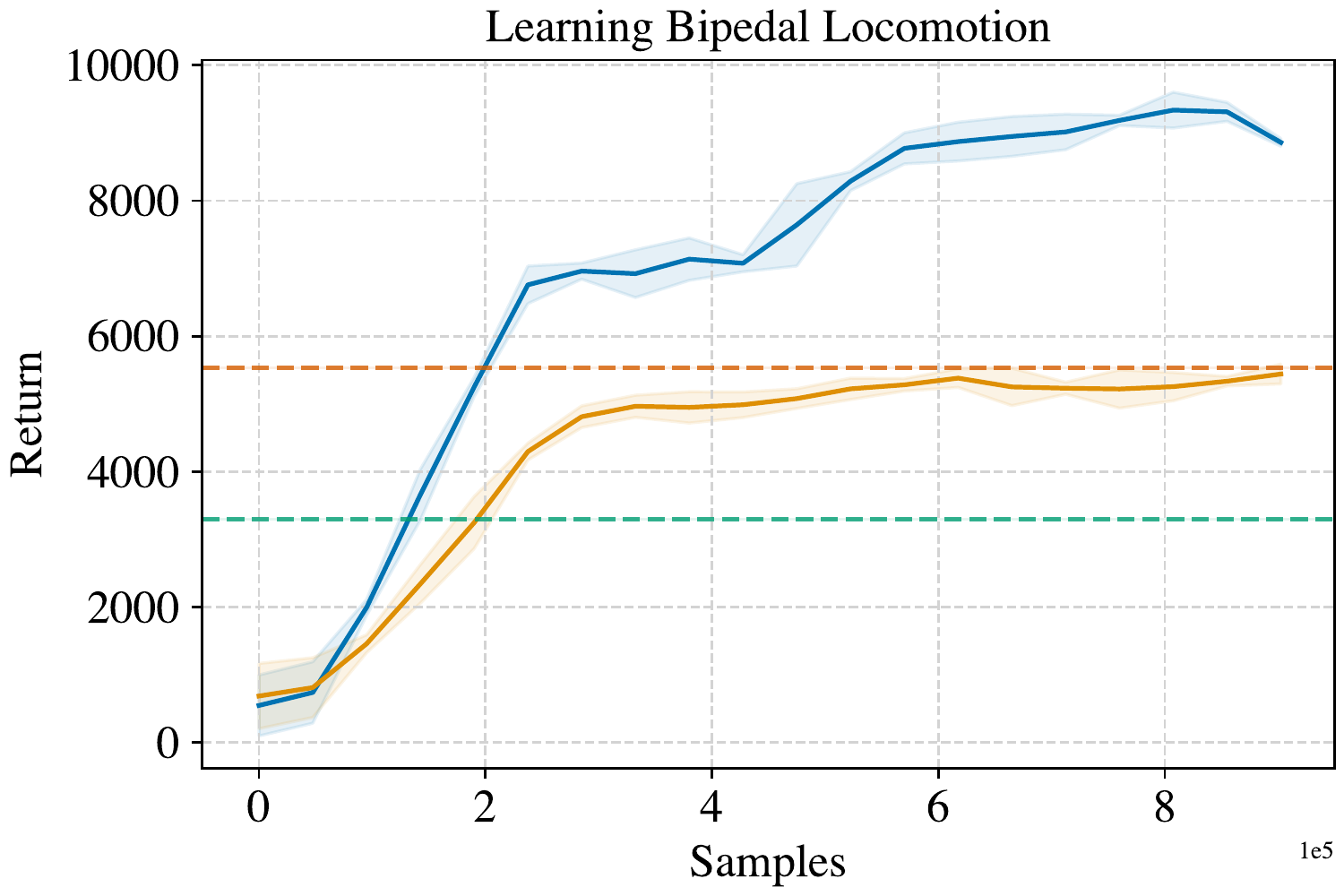}
\caption{
\footnotesize Learning curves for the jumping (left) and bipedal navigation (right) tasks comparing learning from scratch (solid yellow curve) to~\metabbr (solid blue curve) using data from a source policy (dotted green line), shown to 1M steps. Since the policies from scratch had not yet converged for the jumping task, we gave an additional 2 million steps and visualized the average at convergence as well (yellow dotted line); we omit its variance for clarity. We see that in both cases~\metabbr is able to far surpass the policies learned from scratch and the source policies.
}
\label{fig:plot-curriculum}
\end{figure}
To answer (1), we first try to train a policy from scratch on the jumping and bipedal locomotion tasks we defined in~\autoref{sec:applications} (using the same underlying policy optimizer as~\metabbr). We found that we were unable to train a policy to solve either task in this way (see the yellow curves in~\autoref{fig:plot-curriculum}). For the jumping task (the left plot),~\metabbr (blue) consistently achieves $\mathbf{2\times}$ the return achieved by training for the task from scratch, which in addition to its poor average has \emph{very} high variance across seeds. We found that the policies either could not discover the behavior necessary to jump over the hurdles, or learned to hobble over the hurdles but in a very unnatural manner (both behaviors pictured in~\autoref{fig:jumping-qualitative}, blue). 
As shown in the second plot of \autoref{fig:plot-curriculum}, learning the bipedal navigation task from scratch consistently converges to a local optimum. The robot learns to `sit' while facing the goal (see ~\autoref{tab:real-biped-quantitative}), as this allows the robot to collect reward for its root orientation without having to accomplish the much harder task of getting up and walking.
These experiments confirm that it is indeed unlikely for a robot trained from scratch to perform a complex task to acquire desirable behaviors.

\begin{figure}[t]
\includegraphics[width=\linewidth]{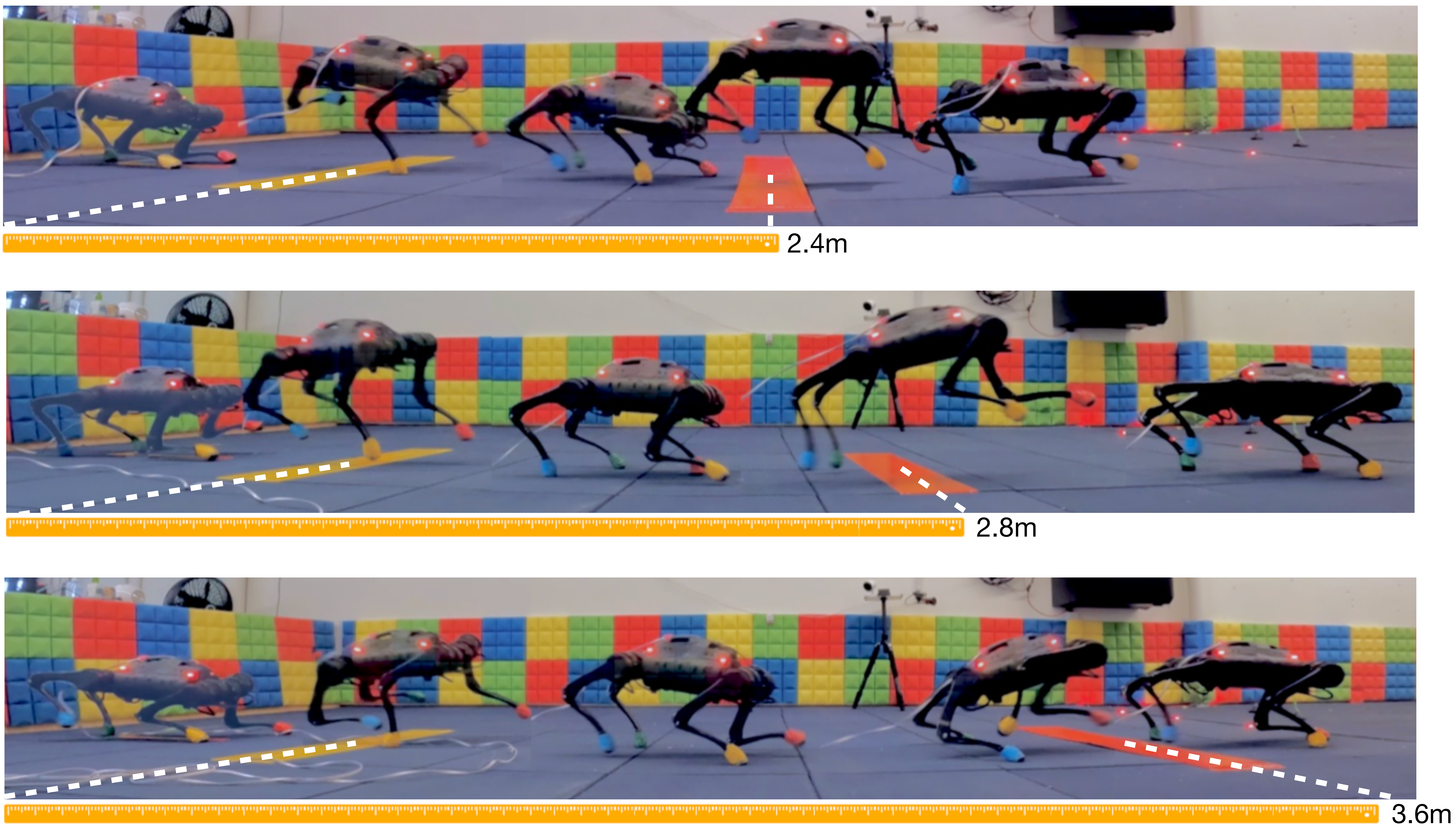}
\caption{\footnotesize Example successful rollouts from evaluating our jumping policy in the real world with different spacings of hurdles. From top to bottom the spacing is at 2.4m, 2.8m, and 3.6m, with success rates over 8 trials of \textbf{75\%, 100\%, and 75\%}, respectively. We see that our policy exhibits the desired behavior (derived from the motion imitation policy) while being robust to task variation, but furthermore, it is able to be deployed on hardware. Videos of all evaluation trials can be found in the supplementary material.}
\label{fig:real-jump-qualitative}
\end{figure}

\begin{figure*}[!ht]
\begin{subfigure}{0.35\linewidth}
\includegraphics[width=\linewidth]{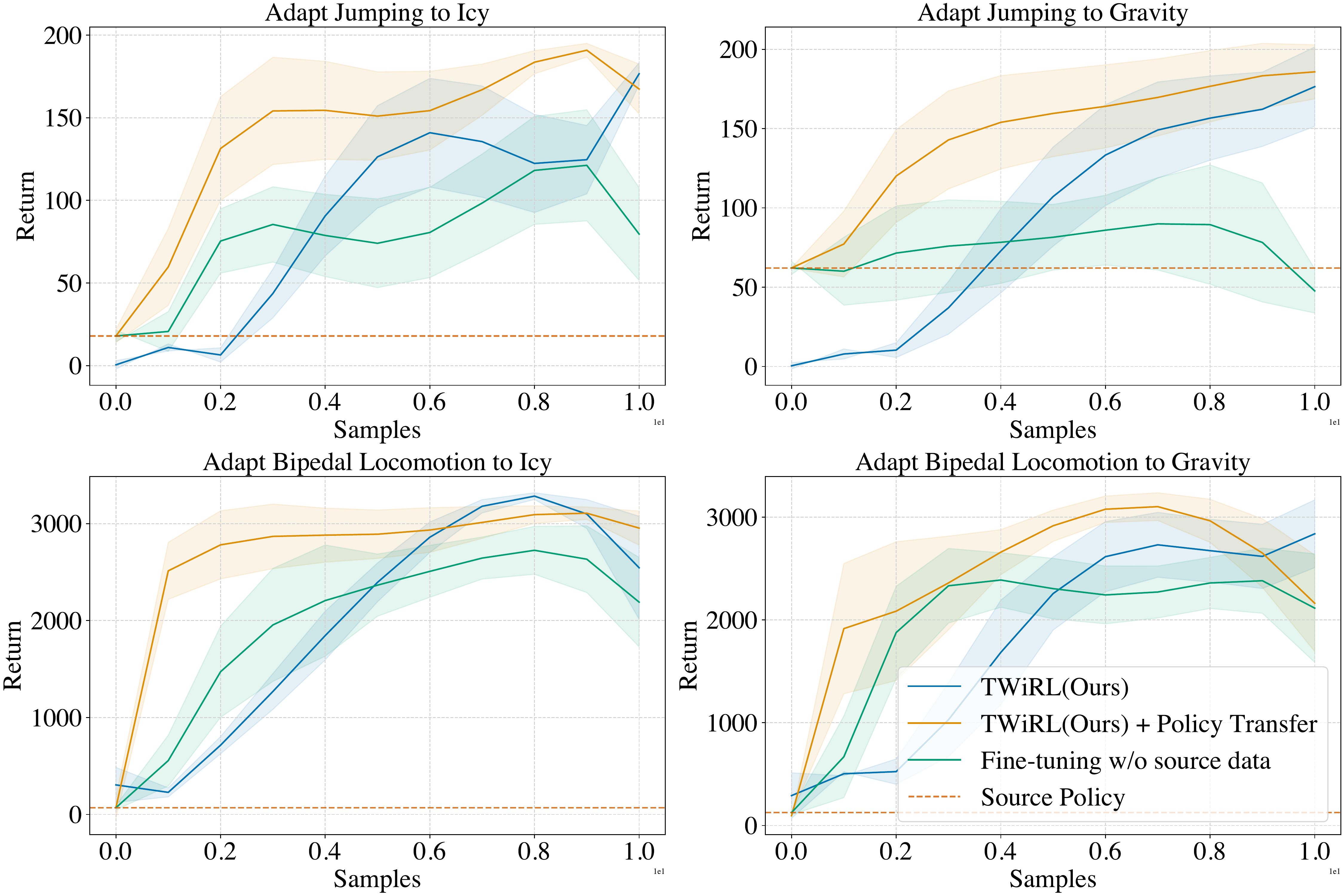}
\caption{\footnotesize Learning curves for adapting the jumping (top) and bipedal navigation (bottom) tasks to the low friction (left) and low gravity (right) environments. }
\label{fig:dynamics-learningcurves}
\end{subfigure}
\hfill
\begin{subfigure}{0.62\linewidth}
\includegraphics[width=\linewidth]{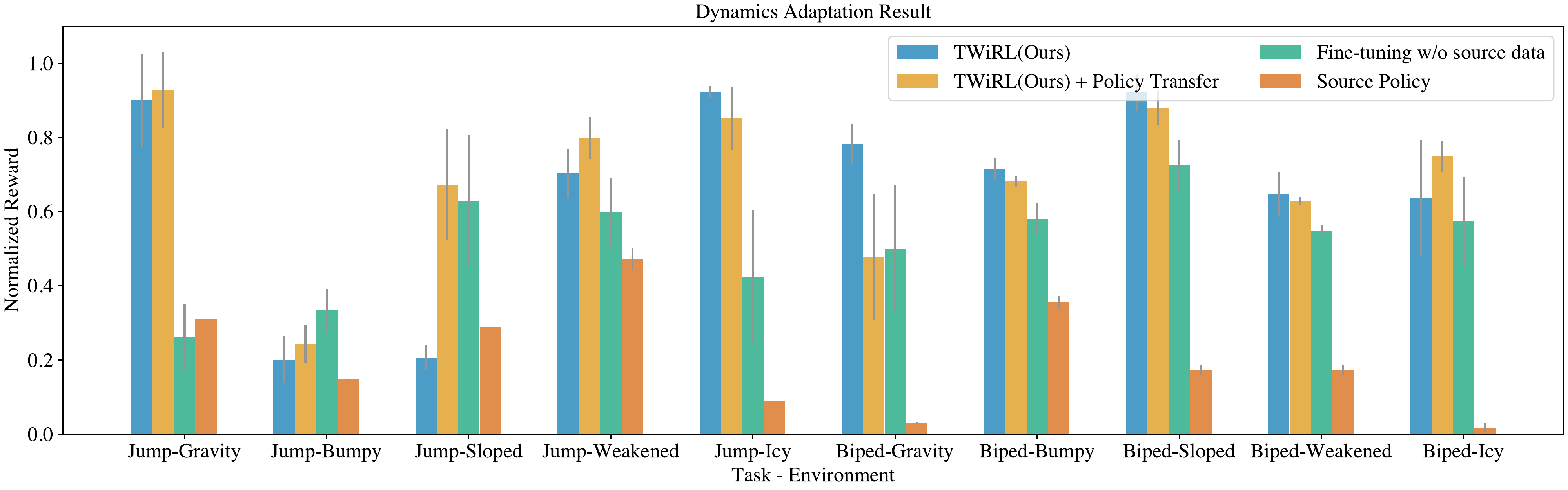}
\caption{\footnotesize 
We report the performance for each task and environment combination after 100k samples' worth of fine-tuning with variants of~\metabbr (meaning transferring the source policy data, blue and yellow) compared to discarding the data (green). For reference, we also report the zero-shot performance of the source policy in the target environment (orange).
}
\end{subfigure}
\caption{\footnotesize Evaluating~\metabbr in adapting the jumping and bipedal navigation skills to 5 different environments (see~\autoref{sec:experimentsetup}) given a budget of 100k samples---corresponding to roughly 1.5 hours real-world time without any overhead. On the left, we show the learning curves for two environments in order to visualize the learning speed/dynamics and on the right, we summarize results for all environments. We see that in all cases, a variant of~\metabbr is able to successfully allow the policies to adapt to new environments. Please refer to the supplementary material for all learning curves.}
\label{fig:dynamics-quantitative}
\vspace{-.5cm}
\end{figure*}
To answer (2), we test whether using \metabbr to incorporate data from policies trained with different objectives effectively overcomes the challenges that prevent successful learning from scratch. As discussed in~\autoref{sec:experimentsetup}, for jumping, we use a policy trained to imitate an animated leaping motion, and for bipedal navigation, a policy trained solely to stand up on its hind legs. For both tasks, the corresponding source policy is suboptimal when evaluated with respect to the target task---in the jumping task we consider variable spacing between hurdles, so following a prescribed trajectory will fail; in the bipedal navigation task, we consider a goal-conditioned task, so the one trained without a notion of this goal will not work. In fact, we quantify the quality of source policies with the dashed green lines in~\autoref{fig:plot-curriculum} (averaged over 100 trials). However, we expect these behaviors to nonetheless provide useful information for learning the target task.  We see that by incorporating this suboptimal data,~\metabbr (solid blue curve) converges to an effective policy for both tasks in fewer than 1 million samples. Interestingly, we find that the resulting policy empirically also retains the naturalistic style exhibited by the source policy (see~\autoref{fig:jumping-qualitative} and~\autoref{fig:bipedal-qualitative}). This is not guaranteed since, unlike other works~\cite{Hester2017DeepQF, Rajeswaran2017LearningCD, Rudner2022OnPI, Nair2017OvercomingEI, playbackcontrol, LearningFromDemos}, 
our objective does not specify any explicit requirement for the policy to resemble the source policy data. Instead, we simply provide examples and allow the value-based RL algorithm to discover the strategies that are useful within them. The advantage of this approach is that we give the robot as much freedom as possible to optimize for the desired task. 

For (3), we evaluate our trained policies on a real A1 quadrupedal robot and find that they perform extremely well---we encourage the reader to see the \href{https://sites.google.com/view/rss23dog-agility}{project website} for all videos. 
The jumping policy tends to amble forward, very stably and almost nonchalantly, before \textit{launching} itself over a hurdle. All four feet are simultaneously airborne before it reaches forward with its front legs to touch down, then steadies itself with its hind legs and continues forward. For jumping, we test 3 variations (corresponding to different hurdle placements) of the task, and we visualize the lengths and distances of these hurdles with colored tape on the ground (see~\autoref{fig:real-jump-qualitative}). Our jumping policy is able to consistently jump over both hurdles with variable spacing, land on its feet, and keep running. It does so whilst \emph{carrying its entire body over the full length of the hurdle during one flight phase}. In contrast, the source policy is only able to jump up to one time before face-planting on the real robot, so we relegate these results to the supplementary material. For the bipedal navigation task, we place a goal (indicated by the red `X' on the ground) about a meter from the robot's starting location~(\autoref{fig:real-biped-combined}). Our bipedal policy shoves itself onto its hind legs, walks forward, and approaches its target location, spending on average over 12 seconds upright. The A1 has small rubber spheres for feet, making this feat akin to balancing on tiptoe. The comparison `bipedal' policy trained from scratch never attempts to stand up, only scoots toward the target on knees and one forepaw. Of course, though, the sim-to-real transfer is far from perfect, and we see this as an exciting opportunity for future work. 

Lastly, to answer (4), we conduct experiments fine-tuning policies to 5 different environments listed in~\autoref{sec:experimentsetup}. Since the only shift in this setting is in the transition function (that is, the state space and task remain from the source policy) we repurpose the replay buffer used to train the source policy to comprise $\buff_\src$ rather than having to collect additional data. For the same reason, we are able to test transferring the policy weights here as well. We show the learning curves for both tasks transferred to the simulated ice rink and modified gravity environments in~\autoref{fig:dynamics-learningcurves}. The most general version of our method wherein we do not assume we can transfer policy weights (blue) consistently enables the robot to re-learn the skill in a way that is tailored to the new environment in just 100k samples. Now, as expected, \emph{additionally} transferring the policy weights (yellow) starts off in the new environment with more competency (than with vanilla~\metabbr). We observe that both variants of~\metabbr generally converge to similar performance, but perhaps surprisingly, discarding the policy weights actually sometimes surpasses the policy that was initialized with $\policy_\src$ after 100k samples. More importantly, though, we see that by comparing to a baseline method of transferring the policy and not the experience (green), incorporating the data from $\policy_\src$ makes an essential contribution to good performance.

\section{Discussion and Future Work}

We presented a framework for enabling quadrupedal robots to learn agile locomotion skills, including jumping and walking on the hind legs, by leveraging a transfer learning framework based on incorporating data from prior policies into training. We describe how the same basic transfer learning framework can make it possible to bootstrap more complex agile locomotion skills with simpler or more constrained ones, as well as enable transfer of skills between environments. Our experimental evaluation shows that employing a curriculum with transfer learning can make it more practical to acquire effective policies for jumping and walking on the hind legs to a goal location than training from scratch, and that our proposed approach to transfer learning outperforms more na\"{i}ve alternatives. Finally, we show that the agile skills that our method can learn can be used to control a real-world A1 quadrupedal robot, displaying a high level of agility.

While our work shows various ways to use transfer learning in service to learning more complex skills, it does not prescribe a single recipe that works in all cases---rather, we aim to illustrate a variety of ways in which the same transfer learning framework can be leveraged. This is also a limitation: the two behaviors we show (jumping and hind leg walking) use different curricula, with jumping employing motion imitation to bootstrap hurdle jumping and hind leg walking employing a hand-designed standing reward for pre-training. A more automated framework for curriculum learning that is based off of our approach could be an exciting direction to explore in future work. This could also enable a more diverse range of behaviors to be specified more easily, which could be particularly exciting as it could enable a broad range of agile behaviors for quadrupedal robots.

\section{Acknowledgments}

This work was supported in part by ARO W911NF-21-1-0097, the Office of Naval Research, and Google. Laura Smith is supported by an NSF Graduate Research Fellowship. We thank Philipp Wu, Kevin Zakka, Ademi Adeniji, Dhruv Shah, Qiyang Li, Katie Kang, Dibya Ghosh, and the members of RAIL for their valuable feedback and support.

\balance

\bibliographystyle{unsrtnat}
\bibliography{references}

\newpage
\appendix
\subsection{Experimental Details}

\subsubsection{Further Environment Details}
\paragraph{Observations} We define the state $\state_t$ to be the 2D Cartesian heading to the goal in the global frame, plus a three-step history of the following features: root roll and pitch, roll and pitch velocity, joint angles, root displacement, and previous actions. 
Additionally, the bipedal navigation policy receives 3D vectors representing its root x and z axes in the global frame.
Roll and pitch, root displacement, goal heading, and the axis orientations come from a motion capture system. Root displacement is a 3D Cartesian vector, the position of the robot's torso relative to its position at the beginning of the episode. Roll and pitch velocity come from the robot's internal IMU. For the hurdling policy, the goal in the goal heading is the middle of the next hurdle. When the robot's root position is 40cm past one hurdle, the goal heading begins pointing to the next hurdle.

\paragraph{Actions} The actions are position targets for the 12 joints of the robot. The policy outputs offset $o$ for each joint from a nominal pose, which for each leg is $p=[0.0, 0.9, -1.8]$, at a frequency of 20Hz. We define a range for the offsets so as to obey joint limits as follows: $o_{min}=[-0.6, -1.2, -1.2], o_{max}=[0.6, 1.2, 1.2]$, for each leg. We use a position controller to compute the torques to be applied $\tau = K_p(q^*-q) - K_d\cdot \dot{q}$, where $q^*$ and $q$ are the desired and current joint positions, respectively, and $\dot{q}$ represents the joint velocities. $K_p$ and $K_d$ are the gains and damping for the motors. We find the values for $K_p$ and $K_d$ to be important in learning our tasks, and set them to $K_p = 60$ and $K_d = 3.0$. We smooth the actions from the policy using a low-pass Butterworth filter with frequency cutoff $= 4.0$.

\begin{figure}[h]
\centering
\includegraphics[width=\linewidth]{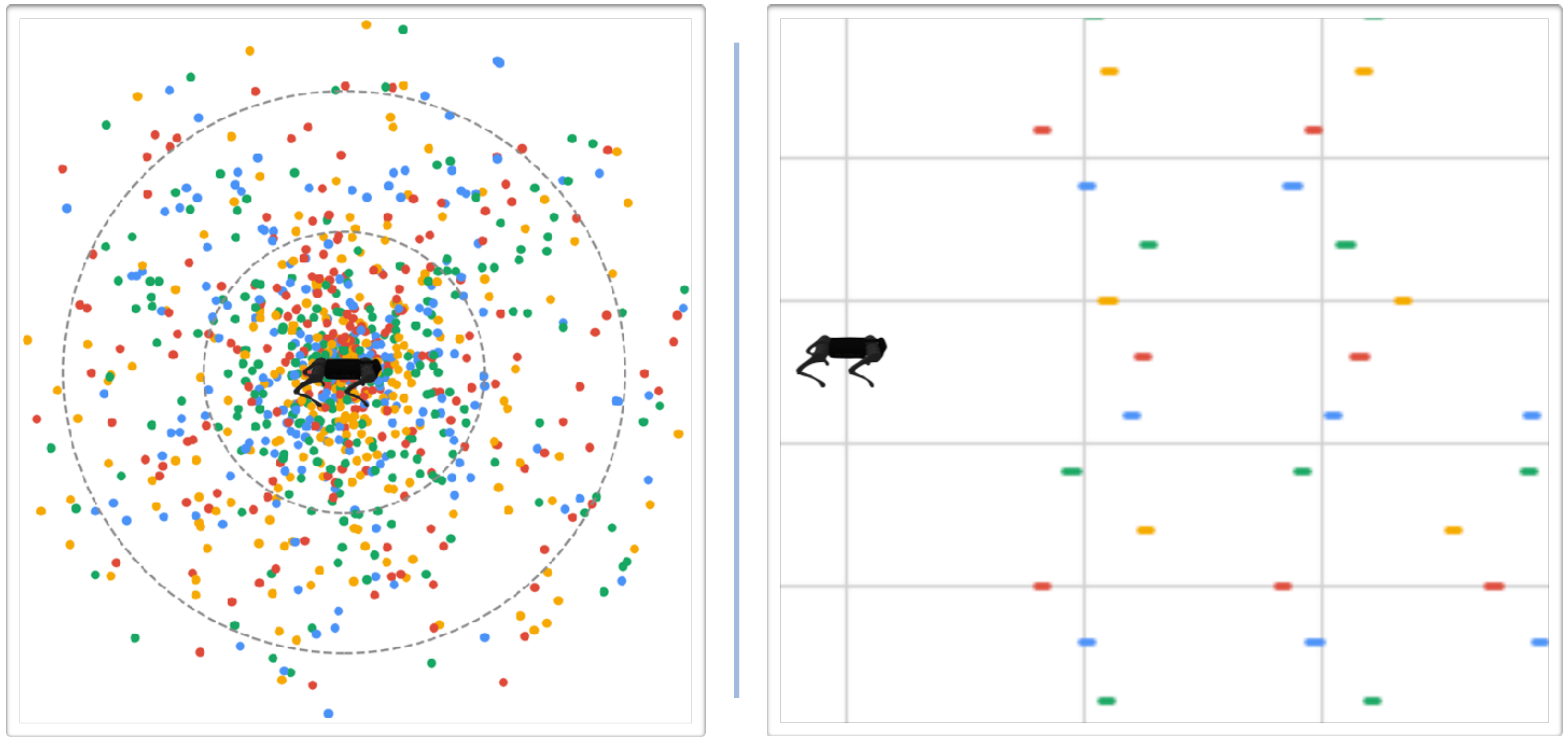} 
\caption{Distribution of goals in the two tasks we consider. (Left) We visualize the bipedal navigation task, where each colored dot is a sample from the goal distribution we defined. The small dotted circle corresponds to a radius of 1m and the larger circle corresponds to a radius of 2m.}
\label{fig:task-distribution}
\end{figure}

\paragraph{Jumping Task} 
We give all details necessary to reproduce our jumping results. First, we detail how we obtained the source policy, then how our task is implemented. 

\noindent\textbf{Training the source policy.}
The jumping source policy is trained to imitate a reference motion clip using the RL framework proposed by~\cite{Peng2020LearningAR}. Given a reference motion $\mathcal{M}$ comprising a sequence of poses, the policy is trained to imitate the motion using a reward function that encourages tracking the target poses at each timestep (see~\autoref{tab:jump-reward}). 
The state $\state_t$ contains a history of 3 timesteps for each of the following features: root orientation, joint angles, and previous actions. The policy also receives a goal $\goal_t$, which comprises the target poses (root position, root rotation, and joint angles) calculated from the reference motion for future timesteps.
In our experiments, we use 4 future target poses, the latest of which is a target for approximately 1 second ahead of the current timestep. 
We adopt the reward function from \cite{Peng2020LearningAR}, where the reward $r_t$ at each timestep is calculated according to:
\begin{equation}
    r_t = \sum_i w^i r^i_t
    \label{eq:jumpeq}
\end{equation}

with $w$'s and $r$'s listed in~\autoref{tab:imitation-reward}.

\begin{table*}[h]
\vspace{-2pt}
        \centering
        \resizebox{.7\linewidth}{!}{%
        \begin{tabular}{l|c|c|c}
            \toprule
            \rowcolor[HTML]{ededed}
            Reward & Formula & Weight & Explanation \\
            \midrule
            Joint pose & $\mathrm{exp}\left[ -5 \sum_j ||q_{\mathsf{ref}}^j - q_{\mathsf{robot}}^j ||^2\right]$ & 0.5 & $q^j =$ 1D local rotation of joint $j$ \\
            Joint velocity & $\mathrm{exp}\left[ -0.1 \sum_j ||\dot{q}_{\mathsf{ref}}^j - \dot{q}_{\mathsf{robot}}^j ||^2\right]$  & 0.05 &   \\
            Root pose & $\mathrm{exp}\left[ -20 ||\mathbf{x}^{\mathsf{root}}_{\mathsf{ref}} - \mathbf{x}^{\mathsf{root}}_{\mathsf{robot}} ||^2 - 10  ||\mathbf{q}^{\mathsf{root}}_{\mathsf{ref}} - \mathbf{q}^{\mathsf{root}}_{\mathsf{robot}} ||^2 \right]$ & 0.15 & $\mathbf{x}^{\mathsf{root}} =$ root's global position, $\mathbf{q}^{\mathsf{root}} =$ root's rotation \\
            Root velocity & $\mathrm{exp}\left[ -2 ||\dot{\mathbf{x}}^{\mathsf{root}}_{\mathsf{ref}} - \dot{\mathbf{x}}^{\mathsf{root}}_{\mathsf{robot}} ||^2 - 0.2  ||\dot{\mathbf{q}}^{\mathsf{root}}_{\mathsf{ref}} - \dot{\mathbf{q}}^{\mathsf{root}}_{\mathsf{robot}} ||^2 \right]$  & 0.1 &   \\
            End effector & $\mathrm{exp}\left[ -40 \sum_e ||\mathbf{x}_{\mathsf{ref}}^e - \mathbf{x}_{\mathsf{robot}}^e ||^2\right]$ & 0.2 & $\mathbf{x}^e =$ robot-relative 3D position of end effector $e$ \\
            \bottomrule
        \end{tabular}
    }%
     \vspace{2pt}
     \caption{Reward terms encouraging tracking a jumping reference motion, from~\cite{Peng2020LearningAR}.}
     \label{tab:imitation-reward}
\end{table*}

\noindent \textbf{Our task.} Jump consecutively over 20cm wide, 1cm tall `hurdles' that are spaced uniformly at random between 1.6 and 2.6 meters apart (see~\autoref{fig:task-distribution}). As such, the robot should run up to each hurdle to jump over it, then land while continuing to run in order to execute the next jump at the right time. We design our reward function with minimal shaping, so as to ensure that it is amenable to adapt to other environments. To ensure that it jumps over the hurdles, we design the reward function to be non-negative and terminate the episode on contact with the hurdle. The reward $r_t$ at each timestep is calculated according to~\autoref{eq:jumpeq} with $w$'s and $r$'s listed in~\autoref{tab:jump-reward}.

\begin{table}[h]
        \centering
        \resizebox{\linewidth}{!}{%
        \begin{tabular}{l|c|c|c}
            \toprule
            \rowcolor[HTML]{ededed}
            Reward & Formula & Weight & Explanation \\
            \midrule
            Forward & 1 if $v_x > 0.5$ else 0.007 & 0.7 & Move toward the next hurdle \\
            Orientation & $\mathrm{exp}\left[ -2 || \mathbf{q}_{\mathsf{root}} ||\right]$  & 0.05 &  Stay upright and pointing forward \\
            Joint velocity & $\mathrm{exp}\left[ -0.1 ||\dot{\mathbf{q}}_{\mathsf{joints}} || \right]$ & 0.15 & Use smoother motions \\
            Hurdle & $\mathsf{count} \left[ \mathsf{hurdles\ passed} \right]$  & 0.1 & Bonus for clearing each hurdle  \\
            \bottomrule
        \end{tabular}
    }%
     \vspace{2pt}
     \caption{Exact definition of each of the reward terms used in our overall jumping reward function defined in~\autoref{eq:jumpeq}.}
     \label{tab:jump-reward}
\end{table}

\paragraph{Bipedal Navigation Task}
Now we give all details necessary to reproduce our bipedal navigation policy. 

\noindent\textbf{Training the source policy.} We define a goal-conditioned navigation task in which the robot must acquire the agility to get up on its hind legs and walk to a desired location while maintaining balance. As such, we first train a policy \emph{just} to get up onto its hind legs by training from scratch with a reward function that sums two terms: 
(i) $\reward^\text{upright}$, a term to make the body perpendicular to the ground and 
(ii) $\reward^\text{max height}$ a term to maximize the height of the robot's CoM. 
For $\reward^\text{upright}$, we first get the robot's forward vector $\texttt{v}_\text{forward}$ by rotating $[1, 0, 0]$ by the robot's orientation. 
We then compute the cosine distance $$\texttt{cos\_dist} = \texttt{v}_\text{forward}^\top [0, 0, 1].$$ 
Lastly, we normalize to be between 0 and 1:
\begin{equation}
\reward^\text{upright} = (0.5 \cdot \texttt{cos\_dist} + 0.5)^2.
\label{eq:normcos}
\end{equation} To encourage the robot to lift itself up, we define
\begin{equation}
\reward^\text{max height} = \exp(\texttt{root\_height}) - 1,
\end{equation} so a height of 0 would correspond to 0 reward and increase with the height exponentially thereafter. The final reward is \begin{equation}
    \reward^\text{stand} = \reward^\text{upright} + \reward^\text{max height}
\end{equation}

We terminate the episode if any part of the robot other than its feet touch the ground.

\noindent\textbf{Our task.} We sample goals uniformly at random along the perimeter of a circle, whose radius is sampled from an exponential distribution with mean 3/2, so as to train the robot to walk to a wide variety of goal locations (see left of~\autoref{fig:task-distribution}). The reward function is a combination of the standing reward and a simple forward locomotion reward. Specifically, we define the reward to be
\begin{equation}
    r^\text{stand}_t \cdot (1 + r^\text{facing}_t + r^\text{distance}_t).
\end{equation}

where $r^\text{facing}_t$ follows~\autoref{eq:normcos}, where $\texttt{cos\_dist}$ is computed between the robot's down vector and its global displacement to the goal. Intuitively, this encourages the robot's belly to point directly at the goal (or for it to face the goal). Finally, the distance reward encourages progress towards the goal:
\[
r^\text{distance}_t = 
\begin{cases}
    1, & \text{if } \texttt{curr\_distance} \leq .01 \\
    .5, & \text{if } \texttt{curr\_distance} < \texttt{prev\_distance} \\
    0, & \text{otherwise.}
\end{cases}
\]
where $\texttt{curr\_distance}$ is the $l_2$ distance from the robot to the goal, and $\texttt{prev\_distance}$ is that from the previous timestep.

\paragraph{Transfer to New Environments} The target sim environments that we construct to study shifts in dynamics are:
\begin{enumerate}[(i)]
    \item \textit{Bumpy:} We generate a random heightfield with a maximum height of 5 centimeters. The robot is reset in random positions on the terrain.
    \item \textit{Icy:} We modify the lateral friction coefficient to 0.4.
    \item \textit{Weakened:} We attempt to simulate real-world stochastic motor performance dynamics by randomly weakening legs and shifting motor temperatures.
    \item \textit{Sloped:} We place the robot next to and point it towards an 8-degree incline.
    \item \textit{Gravity:} We simulate low-gravity environments by modifying the gravity to 3.7 $m/s^2$. We found that the gravity on the moon (1.62 $m/s^2$) was not amenable to the jumping task, so we raised it so as to allow the robot to eventually settle.
\end{enumerate}
Since for some of the domains, the task only makes sense for certain goals, we do not randomly sample goals for the dynamics fine-tuning task. For example, to contend with the dynamics shift imposed by the incline, the goal needs to be placed \emph{on} the hill.

\subsubsection{Training} 

We use the open-sourced JAX implementation of the DroQ algorithm (\url{https://github.com/ikostrikov/walk_in_the_park}) with a full list of hyperparameters in~\autoref{tab:hyperparameters}. 

\begin{table}[h]
\centering
\begin{tabular}{l|c}
\toprule
\rowcolor[HTML]{ededed}
Parameter                    & Value   \\ \midrule
Batch size                            & 256                   \\
Discount ($\gamma$)                             & 0.99              \\
Optimizer                             & Adam              \\
Learning rate                             & $3\times 10^{-4}$              \\
Critic EMA weight ($\rho$)          & 0.005 \\
UTD ratio   &  20 \\
Network Width                        & 256 Units \\
Network Depth & 2 Layers \\
Initial Entropy Temperature ($\alpha$) & 1.0 \\
Target Entropy & $-\dim(\mathcal{A}) / 2$ \\

\midrule
\multicolumn{2}{c}{\metabbr Hyperparameters} \\
\midrule
Source sampling ratio ($\phi$) & 0.5 \\
\bottomrule
\end{tabular}
\caption{Hyperparameters shared for training source policies with DroQ and for transferring with~\autoref{alg:finetune}.}
\label{tab:hyperparameters}
\end{table}
\newpage
\begin{figure*}[hpt]
\begin{subfigure}{\textwidth}
\includegraphics[width=.19\linewidth]{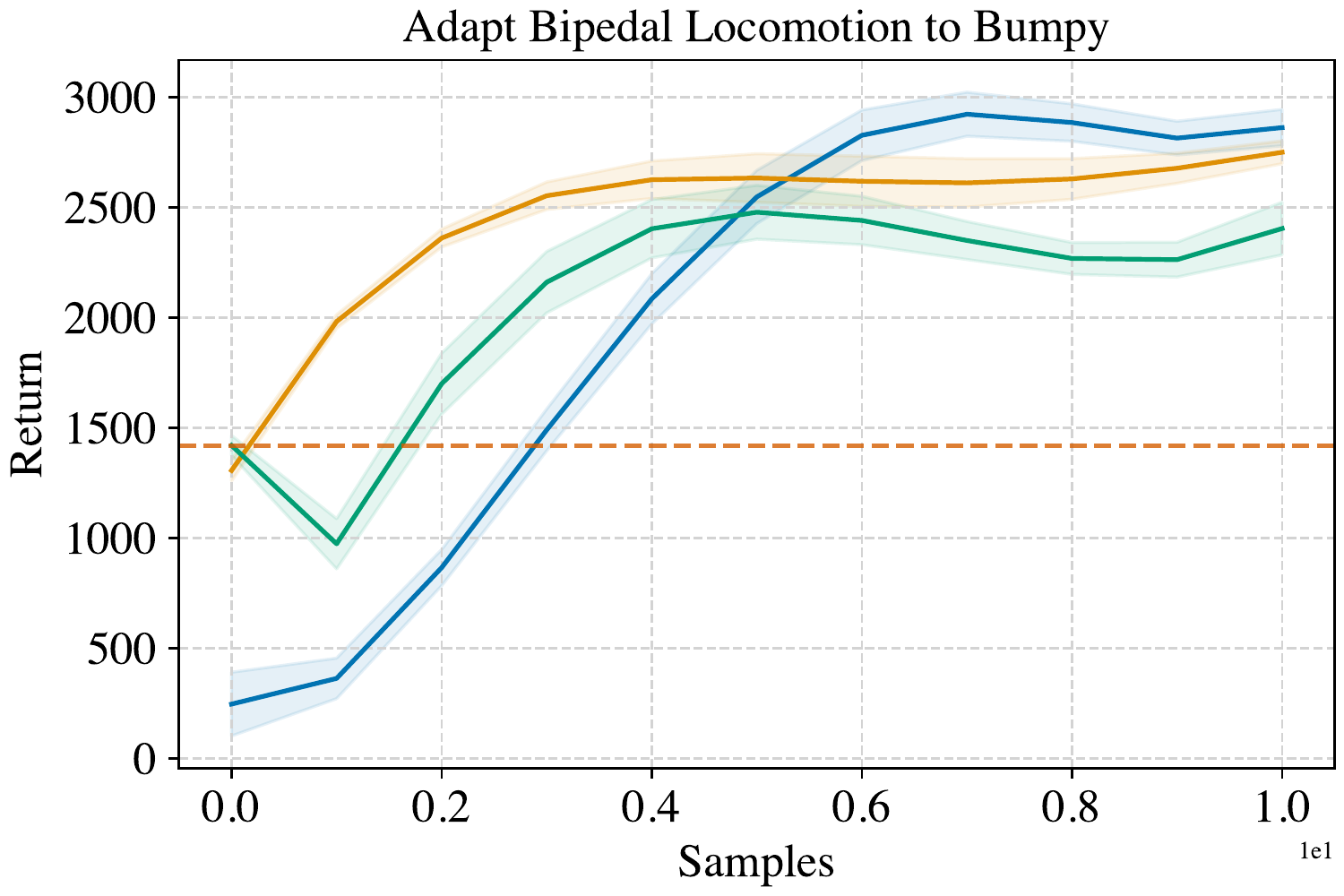}
\includegraphics[width=.19\linewidth]{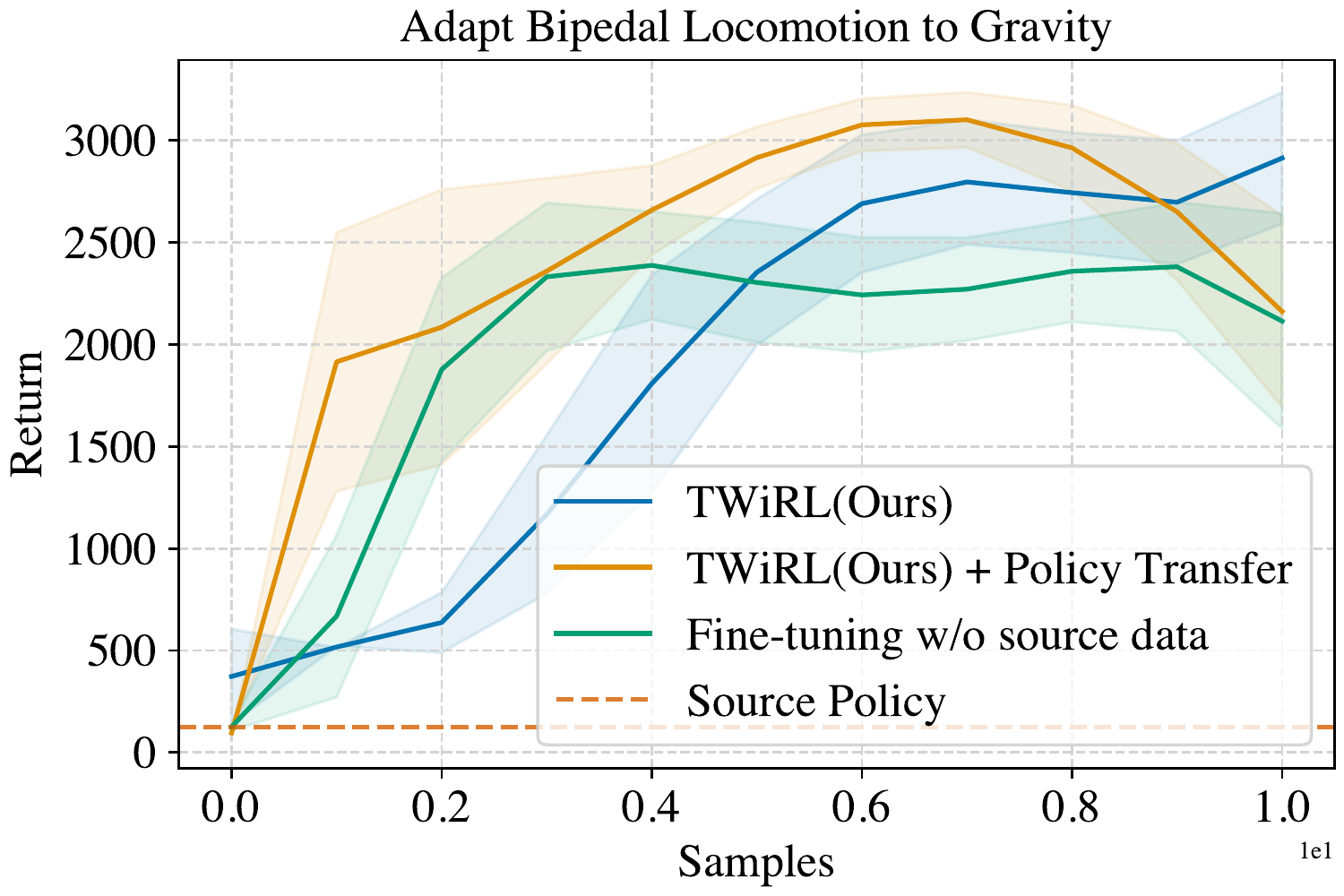}
\includegraphics[width=.19\linewidth]{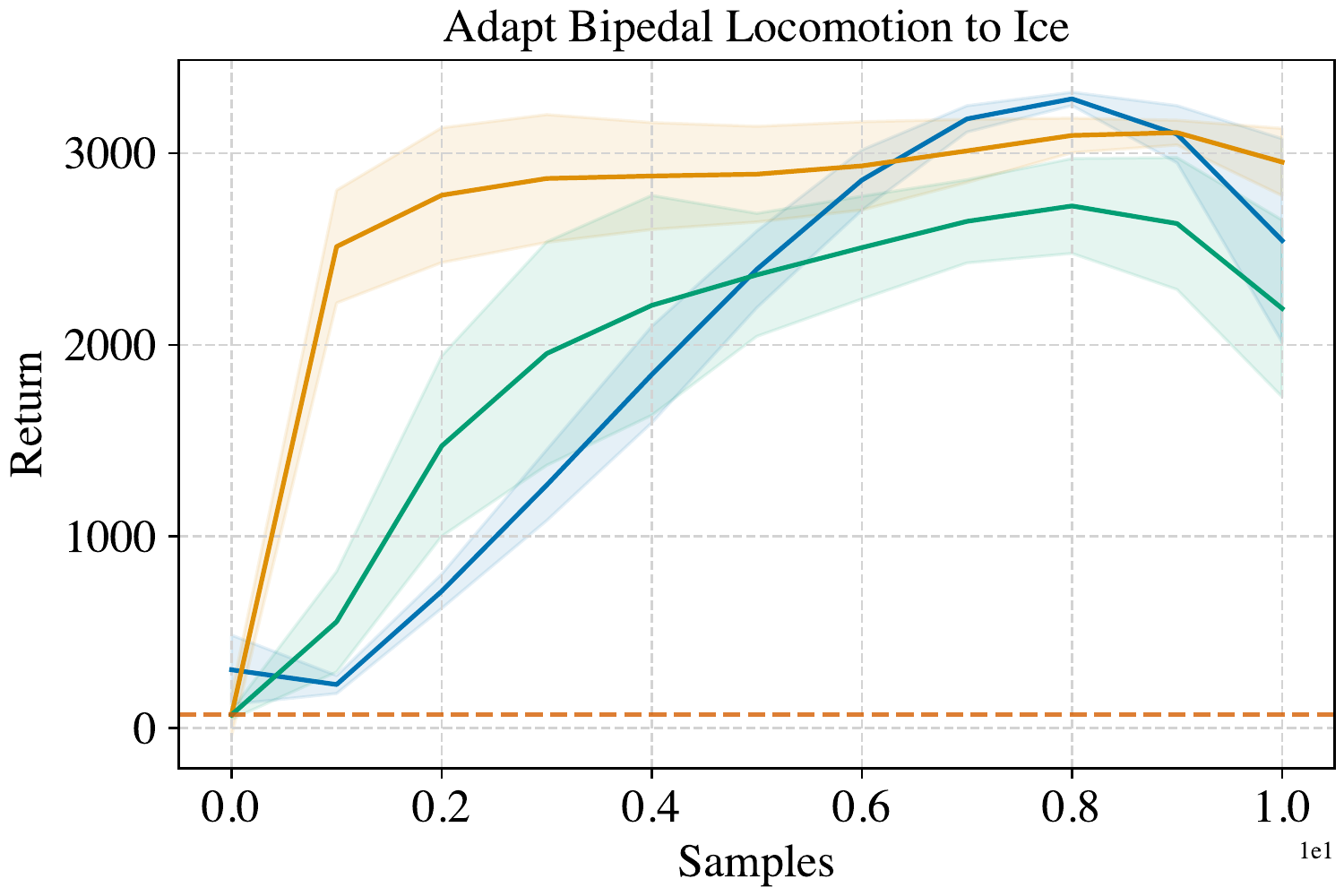}
\includegraphics[width=.19\linewidth]{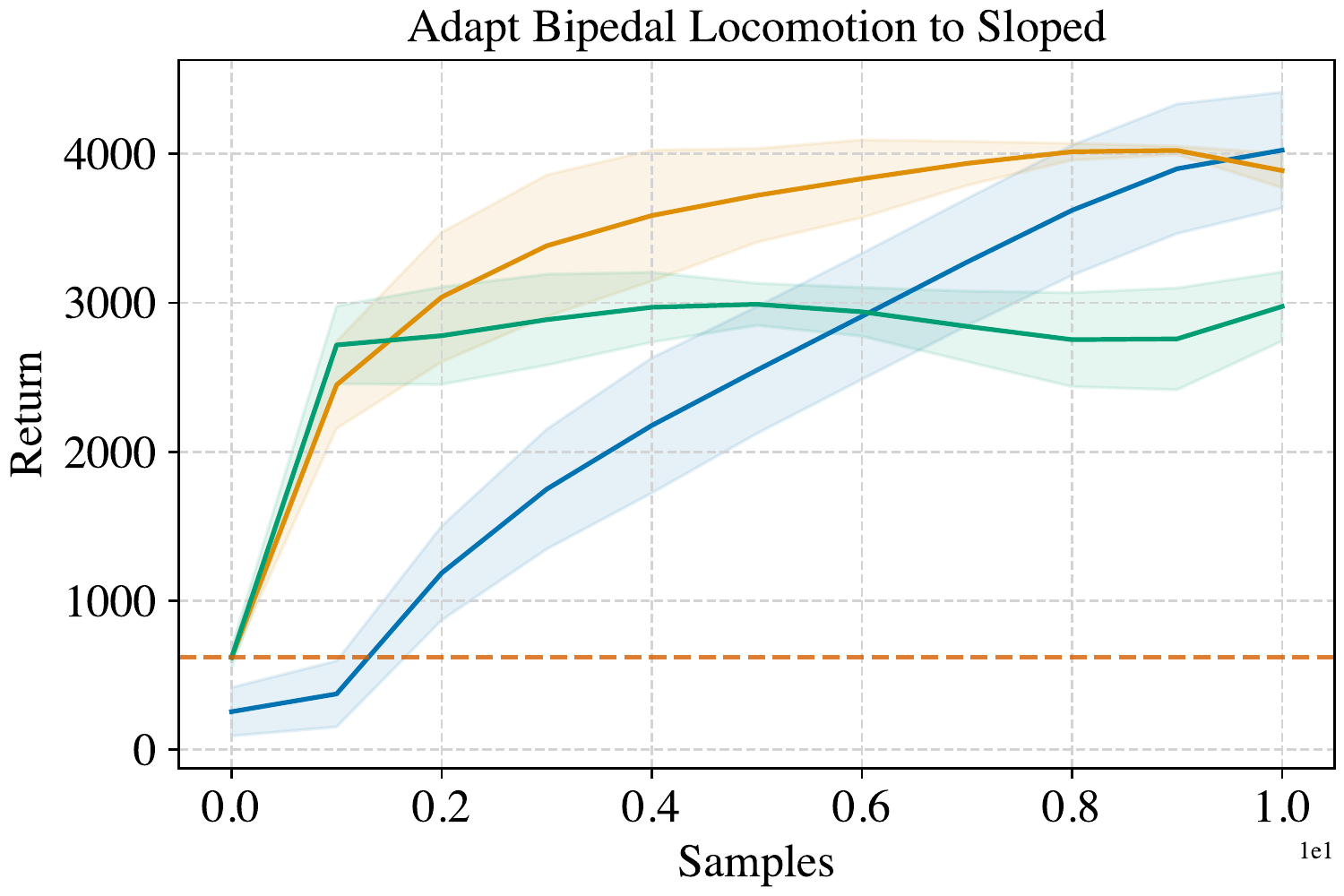}
\includegraphics[width=.19\linewidth]{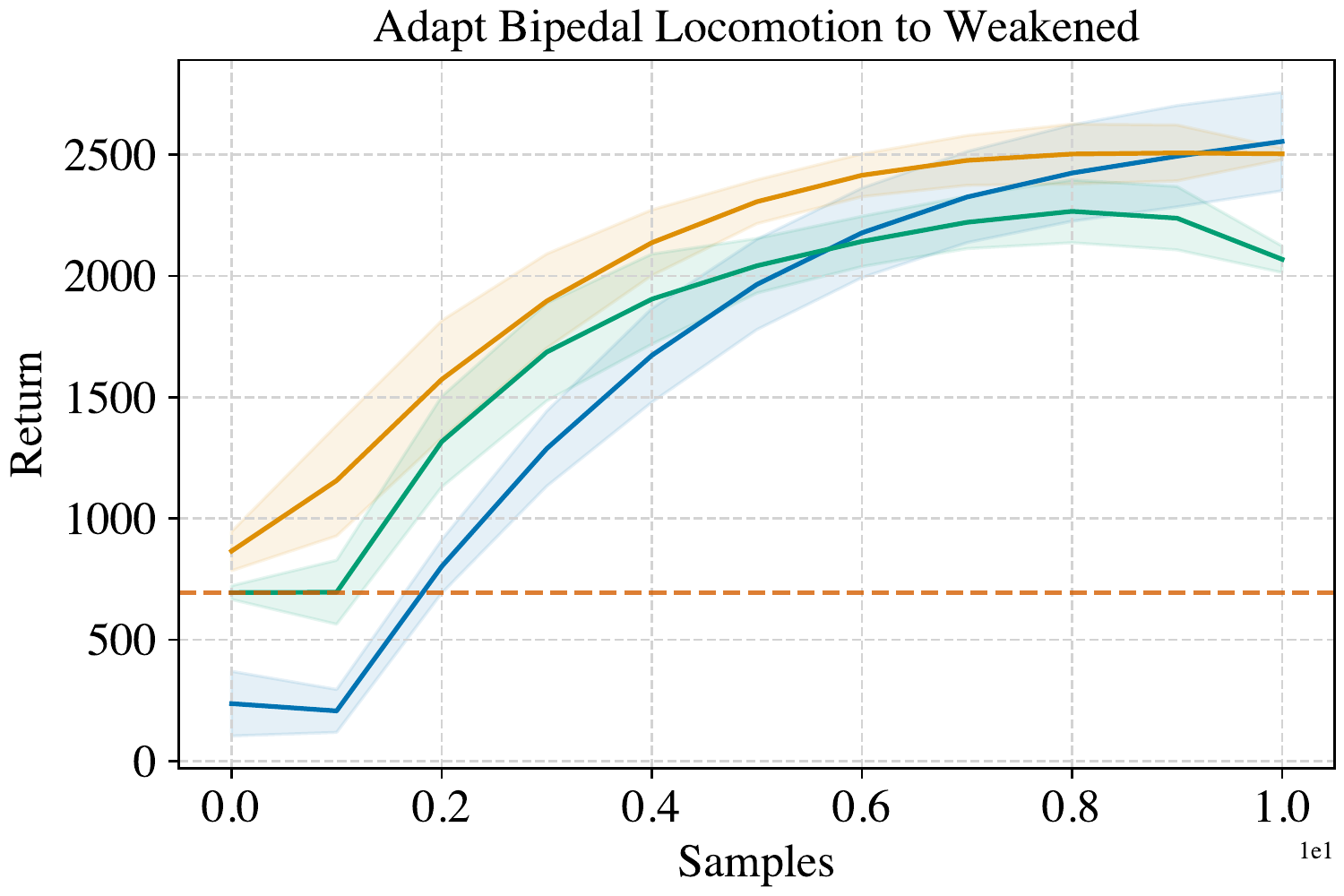}
\caption{Adapting the bipedal navigation policy to the 5 target environments.}
\end{subfigure}%

\medskip
\begin{subfigure}{\textwidth}
\includegraphics[width=.19\linewidth]{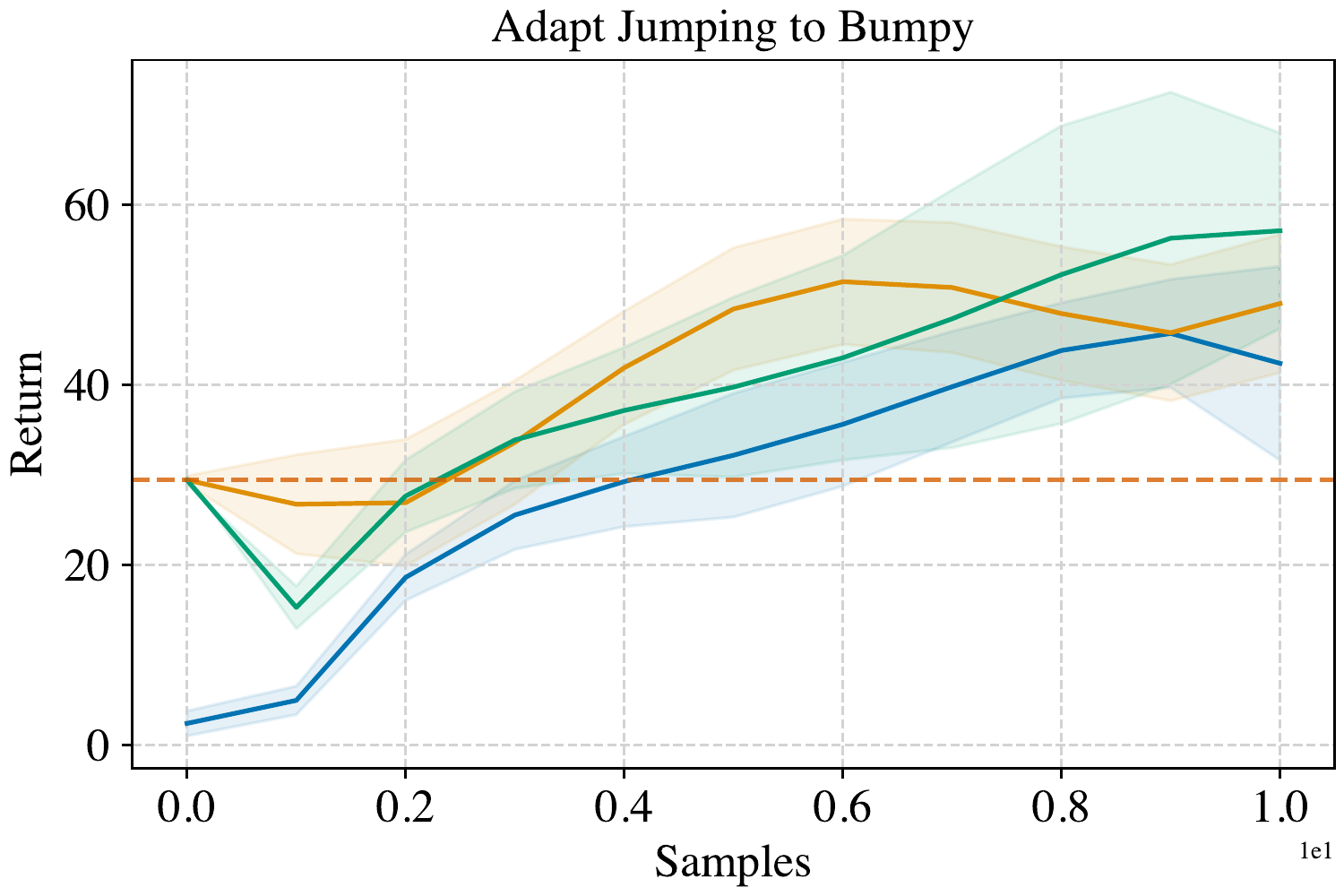}
\includegraphics[width=.19\linewidth]{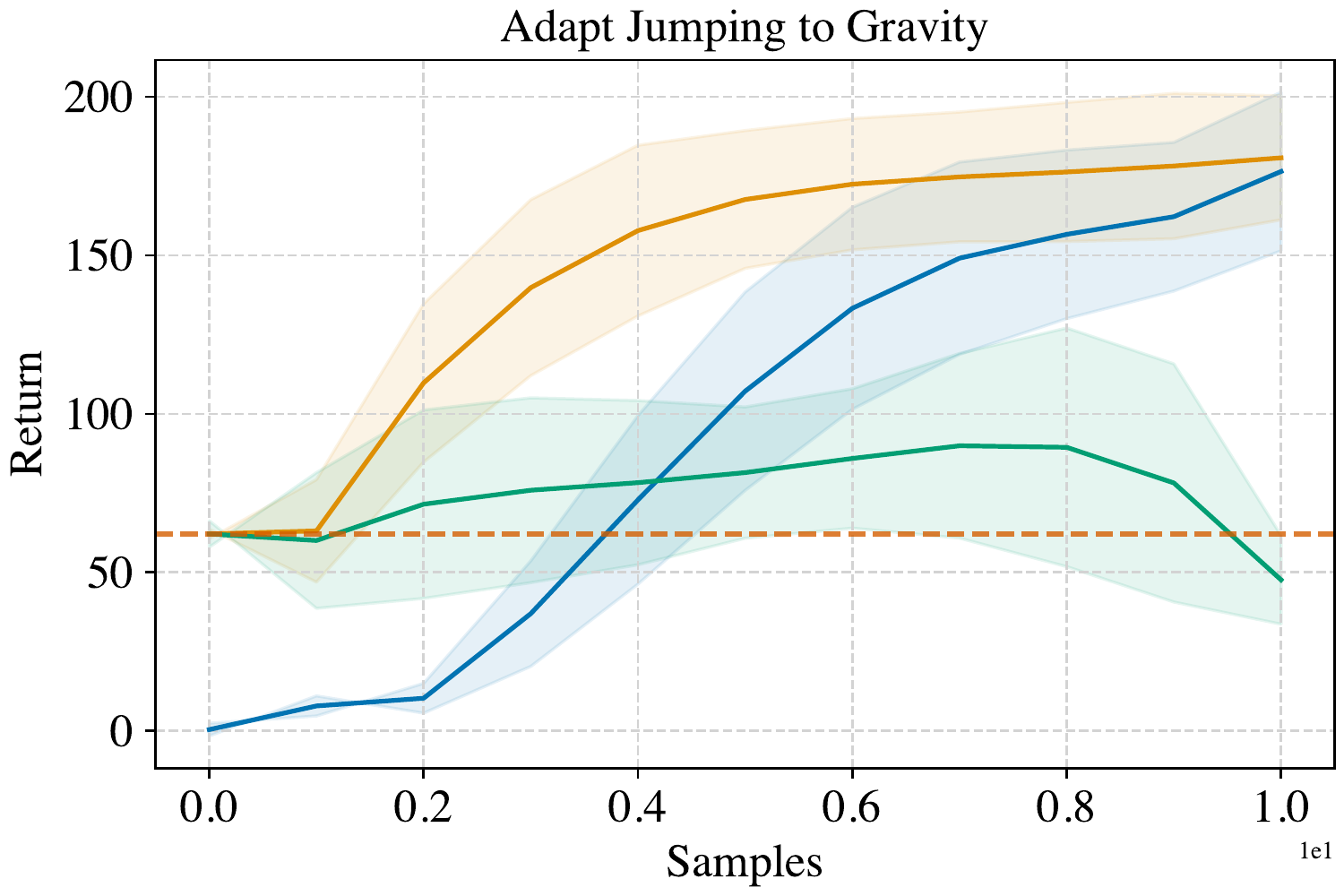}
\includegraphics[width=.19\linewidth]{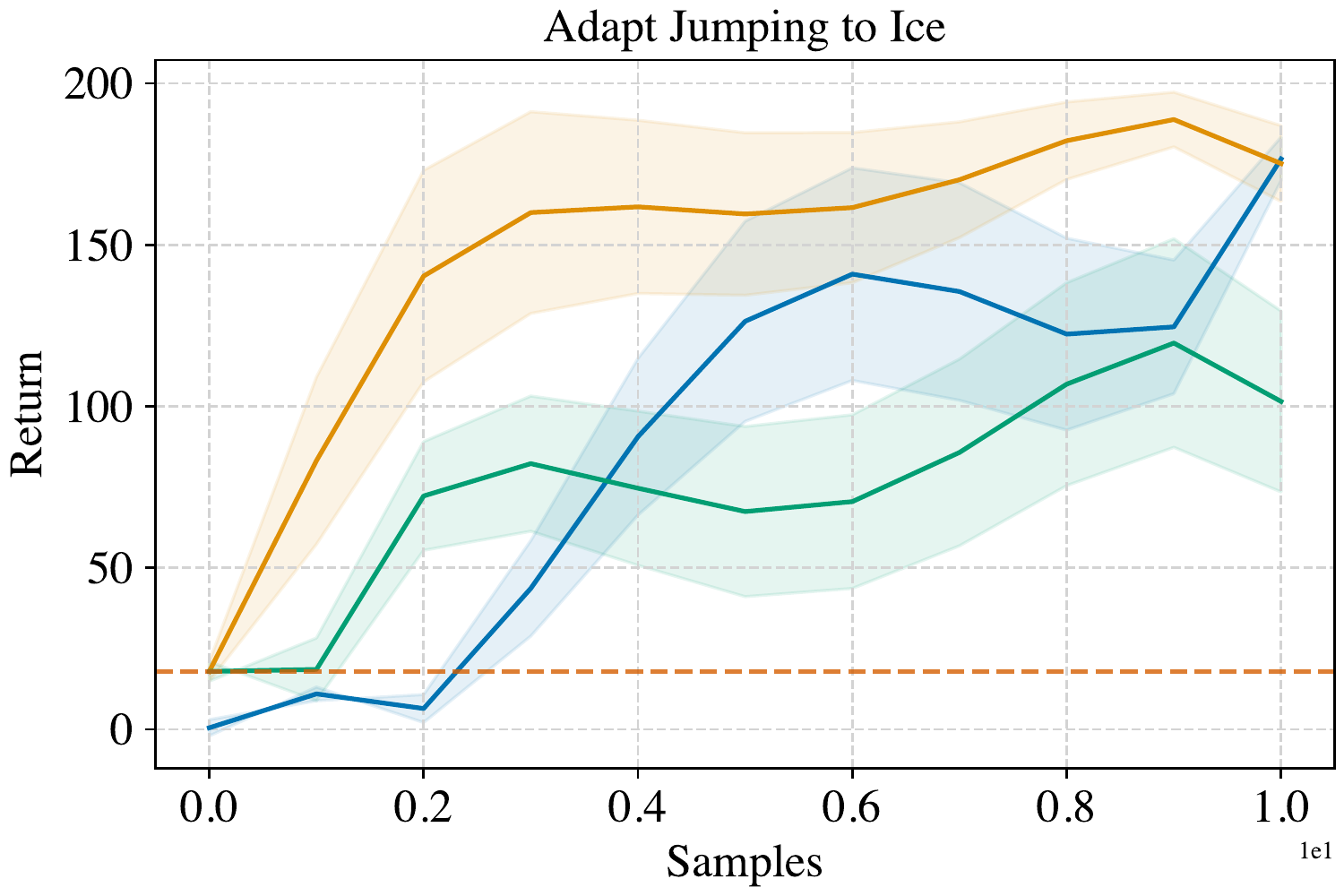}
\includegraphics[width=.19\linewidth]{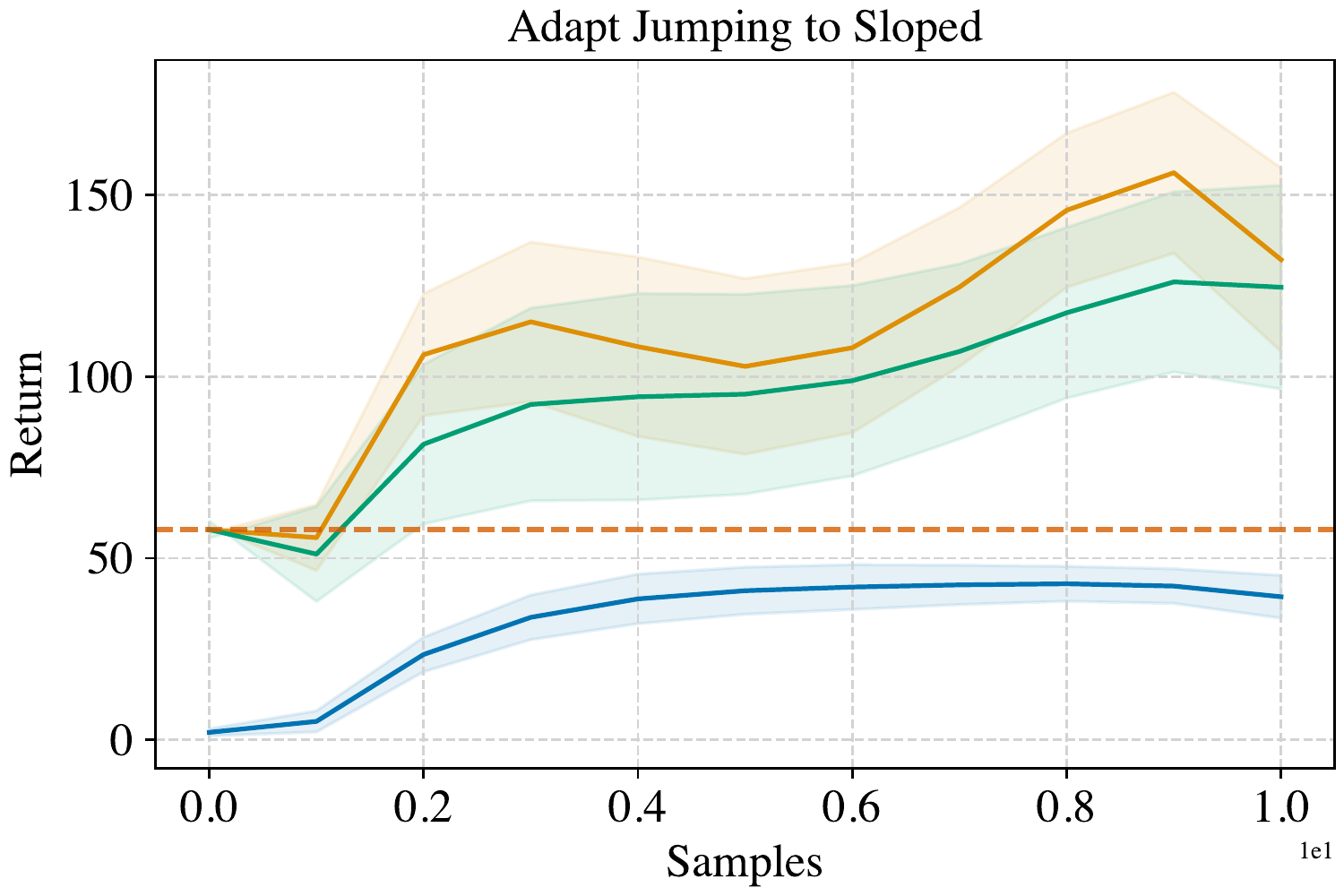}
\includegraphics[width=.19\linewidth]{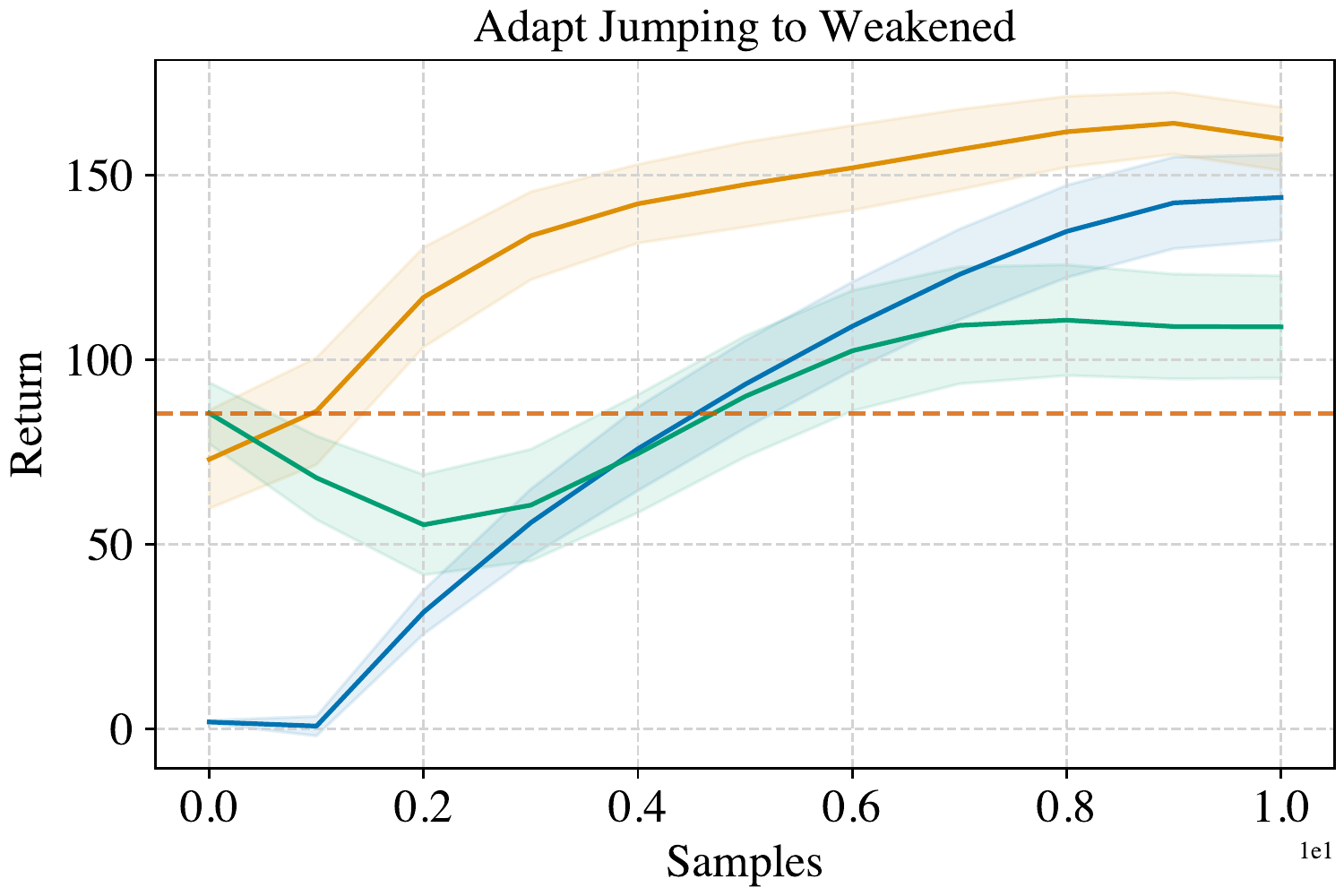}
\caption{Adapting the jumping policy to the 5 target environments.}
\end{subfigure}%
\caption{Individual learning curves for every task and target environment combination.}
\label{fig:alllearningcurves}
\end{figure*}

\subsection{Detailed Experiments}

\subsubsection{Full results}

We present all learning curves for evaluating~\metabbr's ability to facilitate transfer to different environments in~\autoref{fig:alllearningcurves}. We see that in every case except for adapting the jumping policy to the sloped environment, vanilla~\metabbr successfully learns a policy that works in the new environment. Again, we emphasize that vanilla~\metabbr does \emph{not} start with a pre-trained policy, we simply start with the replay from $\policy_\src$. We see that by transferring the weights works robustly in every case. Lastly, we also emphasize that incorporating the data from the source policy is essential, as we see that training from the pre-trained policy with only data collected in the new environment trains significantly slower than either variant of our method.

\subsubsection{Ablations}
In this section, we conduct a series of ablation studies to evaluate how different design decisions affect the performance of our method. Our focus is on two axes, (i) how much does incorporating \emph{data} from the source policy during training affect learning and (ii) whether using a sample efficient, or using a high \emph{update-to-data} ratio, algorithm to efficiently incorporate the data matters. For these experiments, we use dynamics transfer as our goal with a budget of 100k samples for fine-tuning as we can then test \emph{multiple} target environments. 

\begin{figure*}[t]
\includegraphics[width=.95\linewidth]{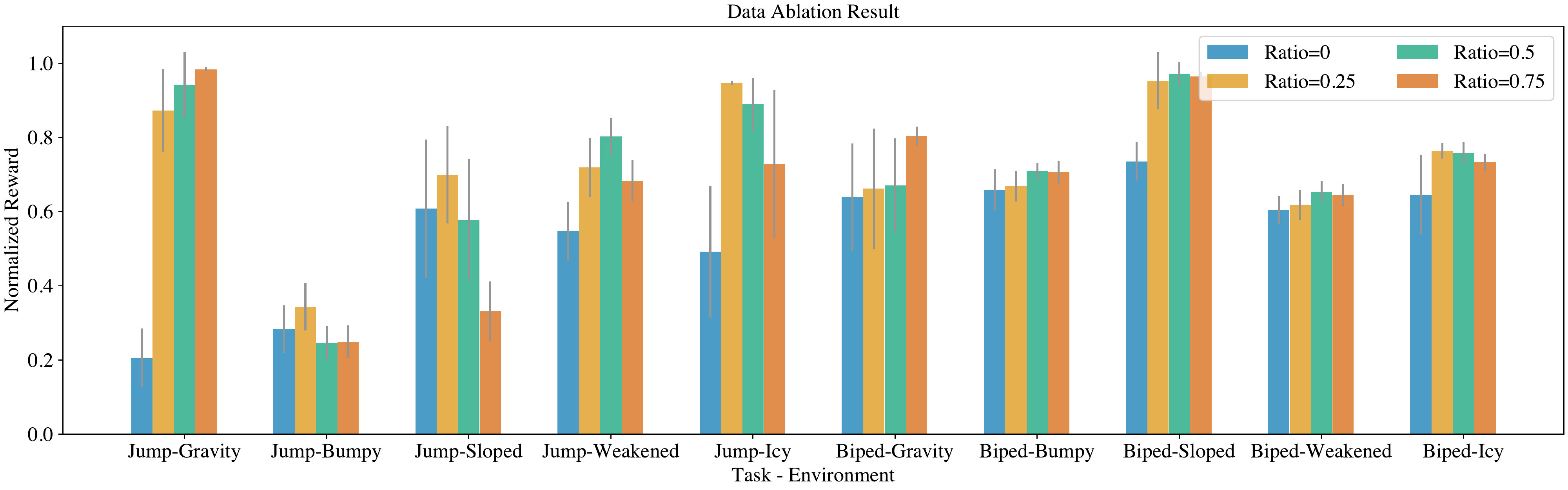}
\caption{Studying the effect of using data from $\policy_\src$.}
\label{fig:plot-data-ablation}
\end{figure*}

\begin{figure*}
\includegraphics[width=.95\linewidth]{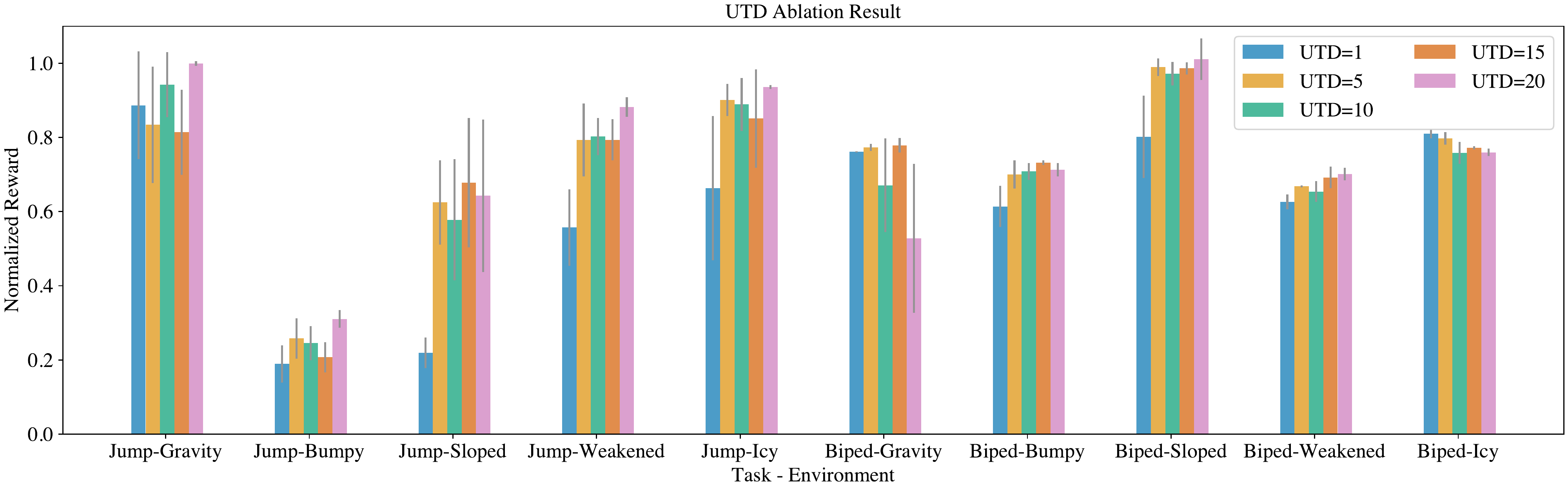}
\caption{Studying the effect of using high UTD algorithms for incorporating off-policy data. We find that using an UTD ratio > 1 is important for stable learning particularly when learning the jumping policies in a new environment.}
\label{fig:plot-utd-ablation}
\end{figure*}

\paragraph{Data ablation}
Our method updates the policy  $\policy_\theta$ using data from both the offline source buffer $\buff_\src$ and an online replay buffer $\buff_\tar$ with a fixed ratio $\phi$, with the procedure defined in~\autoref{alg:finetune}. Here we test 4 ratio options: [0, 0.25, 0.5, 0.75], where 0 indicates that we \emph{only sample data from the online replay buffer}, meaning we do not incorporate the source policy data at all. $\phi=0.5$, which we end up using in our main experiments, indicates we sample half the data from the offline buffer and the other half from the online buffer. Results are shown in \ref{fig:plot-data-ablation}.
Intuitively, discarding the pre-training data and sampling entirely from the online experience (corresponding to $\phi=0$) should allow the agent to most accurately fit to the test environment, as its models do not need to fit to heterogeneous data. However, at the start of fine-tuning as the model needs to adapt to data from a different distribution with such little data, we observe this to be very unstable. By incorporate the source buffer, the off-policy RL algorithm exploits the inherent similarities between the source and target environment to improve the policy update. 
The result shows that higher usage of data from source buffer does not lead to higher result since the lack of data from current environment can prohibit the algorithm to understand the property of the target environment, $\phi=0.5$ result with the best performance in most cases. The result also shows that the improvement by incorporating pre-training dataset differs between target environments, some can greatly improve the fine-tuning outcome e.g. icy, weakened, while the progress is limited for some environments. Our assumption is that the difference between the source and target environment can affects the result. We leave the investigation of how to measure the similarity between environment to future work.

\end{document}